\documentclass[10pt,twocolumn,letterpaper]{article}
\usepackage[T1]{fontenc}

\usepackage[pagenumbers]{cvpr} 

\usepackage{cuted} 

\usepackage{currfile} 

\usepackage{caption} 

\usepackage{amsmath,amssymb}
\usepackage{booktabs}
\usepackage{multirow}
\usepackage{algorithm}
\usepackage[noend]{algpseudocode}
\usepackage{tikz}
\usetikzlibrary{positioning,arrows.meta,fit,calc,backgrounds,shapes.misc}
\usepackage{microtype}

\definecolor{cvprblue}{rgb}{0.21,0.49,0.74}
\usepackage[pagebackref,breaklinks,colorlinks,allcolors=cvprblue]{hyperref}

\def\confName{CVPR}
\def\confYear{2027}

\newcommand{\bothC}{\textsc{bothC}}
\newcommand{\minC}{\textsc{minC}}
\newcommand{\meanC}{\textsc{meanC}}
\newcommand{\allC}{\textsc{allC}}
\newcommand{\col}{\textsc{col}}
\newcommand{\sks}{\texttt{sks}}
\newcommand{\Acomb}{A_{\mathrm{c}}}
\newcommand{\Bcomb}{B_{\mathrm{c}}}

\title{MORE: Merge Once, Route Early for Training-Free\\Multi-Subject Composition from Large LoRA Pools}

\author{Tian XIA\\
{\tt\small xia.tian.t4@alumni.tohoku.ac.jp}}

\begin{document}
\maketitle
\begin{strip}
\centering
\vspace{-2.2em}
\includegraphics[width=\textwidth]{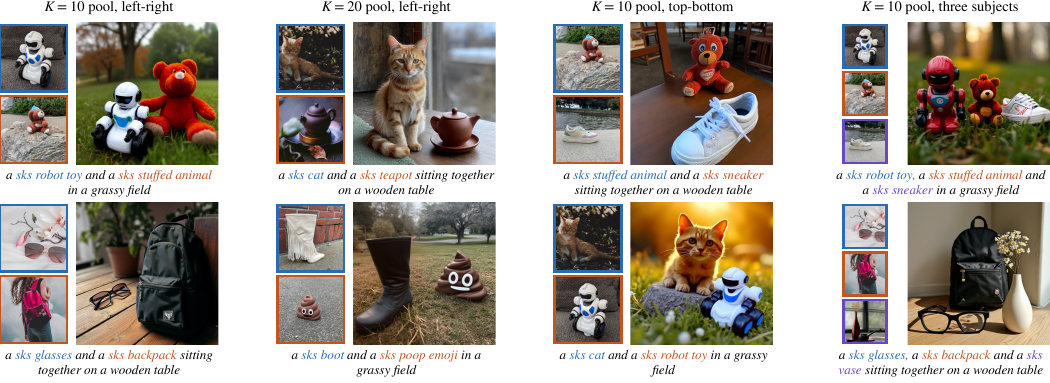}
\vspace{-1.6em}
\captionsetup{hypcap=false}
\captionof{figure}{\textbf{Composing personalized subjects from a pool of $K$ LoRAs in FLUX.} All $K$ subject LoRAs are merged once into a single rank-8 ($K{=}10$) or rank-16 ($K{=}20$) adapter; at inference, the subjects named in the prompt are routed to their regions during the first denoising steps. Each panel: reference images (left), generated image (held-out seed) and its exact prompt, subject words colored as their reference frames; \texttt{sks} is the shared trigger word.}
\captionsetup{hypcap=true}
\label{fig:teaser}
\end{strip}

\begin{abstract}
Public hubs offer many personalized LoRAs per text-to-image backbone, yet composing two of them in one image remains fragile. In FLUX, keeping a pool of $K{=}10$ or $20$ subject LoRAs loaded makes their deltas interfere, and even loading only the two requested LoRAs keeps both subjects in fewer than $4\%$ of images: one subject vanishes, the two fuse, or the image splits into a collage. We present MORE (\emph{Merge Once, Route Early}), a training-free two-stage pipeline for FLUX. \emph{Merge once:} the whole pool is folded into a single fixed-rank adapter by the subspace-regression router of SSR-Merge, made reliable by ridge, rank and calibration choices, and factored exactly into a shared part and per-subject parts. \emph{Route early:} during the first $30\%$ of the denoising steps each subject's part is gated to its layout region; a decaying soft attention bias ties every region to its own subject tokens, and a cyclic swap guidance extrapolates away from the prediction with the subjects' regions permuted, which counteracts region homogenization. On held-out seeds, MORE keeps both subjects in $94.2\%$ ($K{=}10$) and $78.3\%$ ($K{=}20$) of images, against $50.8\%$ / $35.8\%$ for FreeFuse and $25.0\%$ / $24.2\%$ for LoRAShop, and reaches an ImageReward of $1.32$ (best baseline $0.07$), while all $K$ subjects stay in one adapter. With layouts inferred automatically it still keeps both subjects in $70.0\%$ / $55.0\%$ of images. Code is available at \url{https://github.com/sumorday/MORE}.
\end{abstract}

\section{Introduction}
\label{sec:intro}

Low-rank adaptation (LoRA)~\cite{lora} has turned personalization into a commodity: a few images of a pet, a toy or a pair of sneakers yield a small adapter that renders that subject in new scenes~\cite{dreambooth}. A user who owns a \emph{pool} of such adapters naturally asks for several of them at once, ``a \sks{} cat and a \sks{} robot toy in a grassy field''. On large diffusion transformers such as FLUX~\cite{flux}, this request fails in two coupled ways. \emph{Pool crosstalk:} serving any subset of $K$ subjects without reloading means keeping all $K$ deltas active, and since community LoRAs share trigger words and are never trained to stay silent on other subjects, the $K{-}1$ foreign deltas add up: on a subject's own prompt tokens they produce a median $8\times$ ($K{=}10$) and $17\times$ ($K{=}20$) the output energy of the subject's own LoRA. \emph{Coexistence collapse:} even when exactly the two requested LoRAs are loaded, a token-agnostic adapter applies both identities everywhere, and FLUX's joint attention lets every image token read both subjects' text tokens. One subject disappears, the two fuse into a hybrid, or the model escapes into a split-screen collage. Naively loading the two LoRAs keeps both subjects in $1.7\%$ of our held-out images, below the $26.7\%$ of the base model that knows only the class words (\cref{tab:main}); stress tests of multi-subject personalization report the same collapse~\cite{collapse}.

Two lines of work address halves of the problem. \emph{Merging} methods fold many adapters into one weight~\cite{taskarith,ties,dare,regmean,ssr,loraspace}; the data-aware ones, RegMean~\cite{regmean} and its LoRA-subspace form SSR-Merge~\cite{ssr}, keep each LoRA's response on its own inputs and remove most pool crosstalk. But a merged weight is static: it applies to all tokens alike and cannot bind an identity to a location, so merges compose no better than the base model (\bothC{} $0.08$--$0.23$ vs.\ $0.27$ at $K{=}10$, \cref{tab:main}). \emph{Spatial composition} methods route each LoRA to its region with masks: Regional Prompting~\cite{rpflux} with user layouts, FreeFuse~\cite{freefuse} and LoRAShop~\cite{lorashop} with masks inferred from the base model. They assume only the prompt's LoRAs are loaded, apply hard or blended masks, and still lose a subject in about half or more of the images at $K{=}10$ (\cref{tab:main}). Moreover, routing the raw LoRAs is worse than routing a merged adapter (\bothC{} $0.59$ vs.\ $0.94$, \cref{tab:ablation}): routing does not remove the need for a good merge.

\begin{figure*}[t]
\centering
\resizebox{\textwidth}{!}{%
\begin{tikzpicture}[
  font=\small,
  >={Stealth[length=2mm]},
  box/.style={draw, rounded corners=2pt, align=center, inner sep=3pt, fill=white},
  stage/.style={draw=black!40, rounded corners=4pt, fill=black!3, inner sep=5pt},
  lab/.style={font=\scriptsize\itshape, text=black!65, align=center}]
\begin{scope}
  \foreach \k/\c in {0/red!55, 1/orange!70, 2/teal!60, 3/blue!55}{
    \fill[\c] (0.15+0.12*\k, 1.75-0.12*\k) rectangle ++(0.9, 0.55);
    \draw[black!60] (0.15+0.12*\k, 1.75-0.12*\k) rectangle ++(0.9, 0.55);
  }
  \node[font=\footnotesize] at (0.96, 1.62) {$B_kA_k$};
  \node[lab] at (0.95, 0.95) {pool of $K$\\[-1pt] LoRAs};
  \node[box] (cal) at (3.15, 1.85) {calibration\\ $G=\sum zz^{\top}$, $Q$};
  \node[lab] at (3.15, 0.95) {noise latents,\\[-1pt] 5 scenes, $3\times$ prompt tok.};
  \node[box] (rt) at (5.95, 1.85) {router\\ $R=\frac{Q}{n}\big(\frac{G}{n}+\lambda I\big)^{-1}$};
  \node[lab] at (5.95, 0.95) {SSR-Merge router,\\[-1pt] relative ridge};
  \node[box] (svd) at (8.75, 1.85) {rank-$r$ SVD\\ + energy gain $g$};
  \node[lab] at (8.75, 0.95) {$r{=}8$ ($K{=}10$),\\[-1pt] $r{=}16$ ($K{=}20$)};
  \draw[->] (1.6, 1.85) -- (cal);
  \draw[->] (cal) -- (rt);
  \draw[->] (rt) -- (svd);
  \node[font=\small, inner sep=1pt] (eq) at (5.0, -0.05) {$\Delta W x \;=\; \underbrace{W_0x}_{\text{shared}} \;+\; \sum_{k=1}^{K} \underbrace{W_k x}_{\text{subject }k}$ \quad (exact, Eq.~\ref{eq:split})};
  \draw[->] ([xshift=9mm]svd.south) |- (eq.east);
  \node[font=\bfseries, anchor=west] at (0.0, 2.75) {Stage 1: merge once \textmd{\textit{(offline, once per pool)}}};
\end{scope}
\begin{scope}[on background layer]
  \node[stage, fit={(0.0,-1.15) (10.15,2.95)}] {};
\end{scope}
\begin{scope}[xshift=10.6cm]
  \node[font=\bfseries, anchor=west] at (0.0, 2.75) {Stage 2: route early \textmd{\textit{(per prompt, training-free)}}};
  \def\sq#1#2#3#4{%
    \fill[#3] (#1,#2) rectangle ++(0.55,1.1); \fill[#4] (#1+0.55,#2) rectangle ++(0.55,1.1);
    \draw[black!60] (#1,#2) rectangle ++(1.1,1.1);}
  \sq{0.2}{1.15}{orange!45}{cyan!35}
  \node[font=\scriptsize] at (0.475,1.7) {$\Omega_1$}; \node[font=\scriptsize] at (1.025,1.7) {$\Omega_2$};
  \node[lab] at (0.75, 0.92) {layout $\Omega$};
  \sq{2.0}{1.15}{cyan!35}{orange!45}
  \node[lab] at (2.55, 0.92) {permuted $\Omega^{\pi}$};
  \node[font=\footnotesize, align=left, anchor=west] at (3.35, 1.72) {$v=v_\theta(x_i;\,y,\Omega)$\\ $v^{\pi}=v_\theta(x_i;\,y,\Omega^{\pi})$ \textcolor{black!60}{(bias only)}\\ $\tilde v = v + w\,M\odot(v-v^{\pi})$};
  \node[font=\footnotesize, align=left, anchor=west] at (3.35, 0.62) {bias: $+p_i$ on $\Omega_j\!\leftrightarrow\!\tau_j$,\; $-\tfrac{p_i}{2}$ on $\Omega_j\!\leftrightarrow\!\Omega_l$};
  \def\u{0.27}
  \fill[orange!18] (0.2, -0.75) rectangle (0.2+8*\u, 0.25);
  \node[font=\scriptsize, align=center] at (0.2+4*\u, -0.58) {gating $W_0{+}W_j$ in $\Omega_j$};
  \node[font=\scriptsize, align=center] at (0.2+18*\u, -0.58) {static merge $W_0{+}\sum_k W_k$};
  \draw[->] (0.2,-0.75) -- (0.2+28.6*\u,-0.75) node[right, font=\scriptsize] {step};
  \foreach \t in {0,8,27}{\draw (0.2+\t*\u,-0.8) -- ++(0,0.1); \node[font=\scriptsize, below] at (0.2+\t*\u,-0.77) {\t};}
  \foreach \t in {0,2,4,6,8}{\fill[blue!70] (0.2+\t*\u+0.5*\u,0.33) circle (1.6pt);}
  \node[font=\scriptsize, text=blue!70, anchor=west] at (0.2+9*\u, 0.35) {swap guidance (5 extra passes)};
  \draw[thick, red!70, domain=0:28, samples=40] plot ({0.2+\x*\u}, {-0.75+0.9*cos(90*\x/28)});
  \node[font=\scriptsize, text=red!70, anchor=west] at (0.2+22*\u, -0.05) {$p_i$};
\end{scope}
\begin{scope}[on background layer]
  \node[stage, fit={(10.6,-1.15) (19.3,2.95)}] {};
\end{scope}
\end{tikzpicture}}
\vspace{-1.6em}
\caption{\textbf{Overview.} Stage~1 merges the pool once with the SSR-Merge router on our calibration statistics, compresses it to rank $r$, and factors it exactly into a shared part $W_0$ and per-subject parts $W_k$. Stage~2 gates the prompt subjects' parts to their regions for the first $30\%$ of the steps, adds a decaying text--region attention bias $p_i$ in all steps, and applies cyclic swap guidance on five early steps; the passive $K{-}|S|$ subjects stay merged at weight one throughout.}
\label{fig:overview}
\vspace{-0.6em}
\end{figure*}
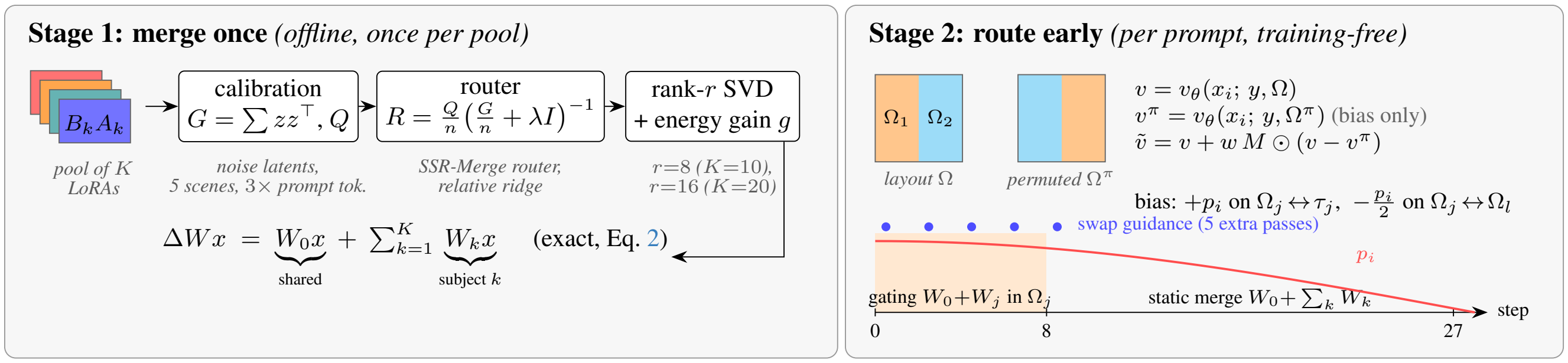

We combine the two in MORE, a training-free pipeline that \emph{merges once and routes early} (\cref{fig:overview}). Stage~1 merges the whole pool into one fixed-rank adapter with the SSR-Merge router, regularized and calibrated so that it holds at $K{=}20$, and factors it exactly into a shared part and one part per subject. Stage~2 reweights these parts per token, but only during the first $30\%$ of the denoising steps, when FLUX commits layout and identity; afterwards the static merge renders details. A soft text--region attention bias, without hard masks, binds each region to its subject's tokens, and a \emph{cyclic swap guidance} contrasts the prediction with one whose subjects are bound to permuted regions, which pushes the regions' features apart instead of letting them homogenize (\cref{fig:mechanism}). Our contributions are:
\begin{itemize}
\setlength\itemsep{0pt}
\item A routable pool merge: the SSR-Merge router with relative ridge, fixed rank and a padding-aware calibration, and an exact factorization of the merged adapter into shared and per-subject parts, so that one rank-$8$/$16$ adapter serves all $K$ subjects (\cref{sec:merge}).
\item Spatio-temporal decoupled routing for MMDiTs: early layout gating of the merged parts, a decaying soft text--region bias, and cyclic swap guidance, a structured instance of degraded-condition guidance that extends to any number of subjects with one extra pass (\cref{sec:route,sec:swap}).
\item A controlled evaluation with held-out seeds, paired bootstrap errors and a collage-aware coexistence metric, against FreeFuse, LoRAShop, Regional Prompting and LoRA Switch/Composite, with ablations of every component and tests on other layouts and on three subjects (\cref{sec:exp}).
\end{itemize}

\section{Related Work}
\label{sec:related}

\paragraph{Merging adapters.}
Weight-space merging combines independently trained deltas without data~\cite{taskarith,ties,dare,tsvm,isoc} or with activation statistics~\cite{fishermerge,regmean,regmeanpp}. For LoRAs, merges operate in the adapters' low-rank subspace~\cite{knots,corespace,lorm}; SSR-Merge~\cite{ssr} solves the RegMean objective restricted to the span of the stacked LoRA factors, and Pico~\cite{pico} and SVC~\cite{svc} damp output directions over-accumulated by many tasks. LoRASpace~\cite{loraspace} is the closest parallel to our pipeline: it computes a pool-wide subspace once and reuses it in a static mode and an input-dependent routing mode. Its routing selects adapters per input, whereas we route the parts of one data-aware merge per \emph{token and step} to bind identities to regions. Our Stage~1 adopts the SSR-Merge router unchanged; we do not claim a new merge rule, only the regularization, rank and calibration choices that make it hold for $K{=}20$ subjects, and an exact factorization that makes it routable. Diffusion-specific concept fusion~\cite{mixofshow,ziplora,klora,orthoadapt,loraclr} typically retrains or constrains the LoRAs, which a user with an existing pool cannot do.

\paragraph{Spatial routing, attention control and guidance.}
Training-free layout control in MMDiTs edits joint attention: Regional Prompting FLUX (RP-FLUX)~\cite{rpflux} masks text--image and image--image attention per region with $-\infty$ entries for the first steps, and related works bind instances by bounded, layered or hard-bound attention or by latent transplantation~\cite{bounded,dreamrenderer,stitch,layerbind,ragdiff,groundit}. TACA~\cite{taca} re-weights cross-modal attention globally with a timestep-dependent temperature; our bias is instead local (each region only towards its own subject's tokens), symmetric and decaying, and adds no hard blocking. For LoRA composition, LoRA Switch and Composite~\cite{multilora} alternate or average per-LoRA predictions, CLoRA~\cite{clora}, OMG~\cite{omg} and LoRA-Composer~\cite{loracomposer} use attention maps or segmenters on UNets, and the FLUX-native FreeFuse~\cite{freefuse} and LoRAShop~\cite{lorashop} infer subject masks from the base model's early attention and route or blend the LoRA outputs. LoRAtorio~\cite{loratorio} weights LoRA outputs per latent patch by their agreement with the base model and supports dynamic selection from a pool; it is the most relevant comparison we could not run. All these methods load only the prompt's LoRAs; we keep the whole pool in one adapter and show that routing a merge beats routing the raw LoRAs. Classifier-free guidance~\cite{cfg} extrapolates away from an unconditional prediction; later work replaces it by a degraded model or condition: perturbed self-attention~\cite{pag}, a weaker model~\cite{autoguidance}, semantically degraded conditions~\cite{cdg}, or normalized attention-space negatives~\cite{nag}. Our swap guidance is a structured instance: the degraded condition keeps prompt, adapter and attention budget and only binds each subject to a wrong region. Multi-subject identity mixing has been measured by detector-based success and collapse rates~\cite{collapse} and by identity-decoupling metrics~\cite{mudi}; our \bothC{} follows the former and adds a collage check.

\section{Method}
\label{sec:method}

\subsection{Problem setting and two kinds of interference}
\label{sec:problem}
A pool holds $K$ subject LoRAs $\{(B_k,A_k)\}_{k=1}^{K}$; LoRA $k$ adds $B_kA_k$, with $B_k\in\mathbb{R}^{d_{\mathrm{out}}\times r_0}$, $A_k\in\mathbb{R}^{r_0\times d_{\mathrm{in}}}$, to each of the $L$ adapted linear layers of FLUX.1-dev~\cite{flux} ($r_0{=}32$, $L{=}342$). Each LoRA is trained DreamBooth-style~\cite{dreambooth} on ``a \sks{} $\langle$class$\rangle$''; the shared trigger is typical of community pools and is itself a source of crosstalk. A request is a prompt $y$ that names $m\ge2$ subjects $S=(s_1,\dots,s_m)$ of the pool, together with disjoint layout regions $\Omega_1,\dots,\Omega_m$ over the $N{=}1024$ image tokens (the $32{\times}32$ latent grid at $512$px); tokens in no region form the background. FLUX processes the joint sequence [text $|$ image] with joint attention in $19$ double and $38$ single blocks, so every image token can attend to every subject's text tokens $\tau_j$ (the span ``\sks{} $\langle$class$_j\rangle$'').

The two failure modes of \cref{sec:intro} have different remedies. \emph{Pool crosstalk} is a property of the weights: keeping all $K$ LoRAs available as one adapter requires each subject's delta to survive on its own inputs despite the $K{-}1$ others, which is what data-aware merging optimizes. \emph{Coexistence collapse} is a property of token routing: a weight shared by all tokens, merged or not, gives every image token both identities, so it cannot be fixed in weight space. We therefore merge once, offline, and route the merged parts per token at inference.

\subsection{Stage 1: calibrated subspace merge}
\label{sec:merge}
Let $\Acomb=[A_1;\dots;A_K]\in\mathbb{R}^{Kr_0\times d_{\mathrm{in}}}$, $\Bcomb=[B_1,\dots,B_K]$, and $z=\Acomb x$ the LoRA-A features of a layer input $x$. Given calibration inputs $\mathcal{X}_k$ of each subject, RegMean~\cite{regmean} asks one delta to reproduce every LoRA on its own inputs, $\min_{\Delta W}\sum_k\sum_{x\in\mathcal{X}_k}\|\Delta W x-B_kA_kx\|^2$. Restricting the solution to the LoRA subspace, $\Delta W=\Bcomb R\Acomb$, and writing $B_kA_kx=\Bcomb E_kz$ with $E_k$ the selector of block $k$, the normal equations give
\begin{equation}
\begin{aligned}
R&=\tfrac{Q}{n}\Big(\tfrac{G}{n}+\lambda I\Big)^{-1},\\
G&=\sum_{k}\sum_{x\in\mathcal{X}_k}\!zz^{\top},\quad Q=\sum_k E_k\!\sum_{x\in\mathcal{X}_k}\!zz^{\top},
\end{aligned}
\label{eq:router}
\end{equation}
where $n$ is the number of calibration tokens. For $\lambda\to0$ this is the router of SSR-Merge~\cite{ssr}, which we adopt; it equals RegMean restricted to the LoRA subspace. We use a ridge relative to the feature scale, $\lambda=\rho\,\mathrm{tr}(G/n)/(Kr_0)$ with $\rho{=}0.05$ ($K{=}10$) or $0.1$ ($K{=}20$), instead of the absolute $\lambda{=}10^{-4}$ of SSR-Merge, so that one value can be shared by layers whose feature scales differ widely.

\paragraph{Calibration.} SSR-Merge computes $G,Q$ from one prompt per subject. With T5's $512$-token context, most of those tokens are padding that looks alike for all subjects, so the regression is dominated by inputs on which no subject should differ. We instead use (i) prompts ``a \sks{} $\langle$class$_k\rangle$ $\langle$scene$\rangle$'' with five scenes disjoint from the evaluation prompts, (ii) noise latents (no training images) at $t\in\{0.1,0.3,0.5,0.7,0.9\}$, and (iii) the statistics of the non-padding prompt tokens added with weight $3$ after per-token normalization. \Cref{tab:ablation} shows that calibration is the largest merge-side factor.

\paragraph{Fixed rank and energy.} With $M=\Bcomb R$, we compute the rank-$r$ truncation of $M\Acomb$ in factored form (thin SVDs of $\Acomb$ and $M$, never forming a $d_{\mathrm{out}}{\times}d_{\mathrm{in}}$ matrix), which gives $\Delta W_r=B_rT\Acomb$ with $B_r\in\mathbb{R}^{d_{\mathrm{out}}\times r}$, $T\in\mathbb{R}^{r\times Kr_0}$. A per-layer scalar $g=(E_{\mathrm{own}}/E_r)^{1/2}$ restores the calibration output energy of the own LoRAs, $E_{\mathrm{own}}=\sum_k\sum_{\mathcal{X}_k}\|B_kA_kx\|^2$, $E_r=\sum_k\sum_{\mathcal{X}_k}\|\Delta W_rx\|^2$. We use $r{=}8$ for $K{=}10$ and $r{=}16$ for $K{=}20$ (a quarter and a half of a single LoRA's rank). Optionally, the few output directions of $\Bcomb$ shared by all subjects (share ${>}0.5$ among the top eight singular directions, ${\approx}25$ of $342$ layers) are damped as in Pico~\cite{pico}; this has no significant effect on composition (\cref{tab:ablation}).

\paragraph{Exact factorization.} Let $P=U_sU_s^{\top}$ project onto those shared directions ($P{=}0$ in most layers) and split $T=[T_1,\dots,T_K]$ into column blocks. Then, exactly,
\begin{equation}
g\,\Delta W_r x=\underbrace{gPB_rT\Acomb x}_{W_0x}+\sum_{k=1}^{K}\underbrace{g(I{-}P)B_rT_kA_kx}_{W_kx}.
\label{eq:split}
\end{equation}
$W_k$ is the merged adapter's response to subject $k$'s own LoRA-A features. Summing all terms recovers the static merge, so any per-token reweighting of the terms is a controlled departure from a well-defined merge; all parts share the output basis $B_r$.

\subsection{Stage 2: spatio-temporal routing}
\label{sec:route}
Stage~2 changes neither weights nor text embeddings. It reweights the terms of \cref{eq:split} per token and step, and adds an attention bias. Subjects not in the prompt are \emph{passive} and keep weight one everywhere, so outside the prompt the adapter is the static merge.

\paragraph{Early layout gating.} For the first $\mathrm{round}(\gamma N_s)$ of the $N_s{=}28$ steps ($\gamma{=}0.3$, steps $0$--$7$), an image token $u\in\Omega_j$ receives $W_0+W_{s_j}$ plus the passive parts, \ie, the partner's part is removed from its region; background tokens receive all parts; text tokens in $\tau_j$ receive only $s_j$'s part, other text tokens all parts. From step $8$ on the gating is off and the full static merge acts on every token. In FLUX the coarse layout and the subjects' identities are decided in the first steps (\cref{fig:mechanism}); later steps refine texture, where the full merge, including the shared part, renders details well. Gating only early keeps \bothC{} and raises \meanC{} over gating all steps (\cref{sec:ablation}).

\paragraph{Soft text--region bias.} In every attention layer and every step we add to the logits, for query $a$ and key $b$,
\begin{equation}
\beta_i(a,b)=\begin{cases}
+p_i & a\in\Omega_j,\,b\in\tau_j\ \text{or}\ a\in\tau_j,\,b\in\Omega_j,\\
-p_i/2 & a\in\Omega_j,\,b\in\Omega_l,\ l\neq j,\\
0 & \text{otherwise},
\end{cases}
\label{eq:bias}
\end{equation}
with $p_i=p_0\cos\!\big(\tfrac{\pi}{2}\tfrac{i}{N_s}\big)$ and $p_0{=}2$. No entry is masked: a region may still attend to the other subject's tokens, only less. This differs from the $-\infty$ masks of RP-FLUX~\cite{rpflux}, which on our merge \emph{reduce} coexistence (\cref{sec:ablation}), and from TACA~\cite{taca}, which rescales all text--image logits uniformly. The image--image term softly discourages regions from copying each other's features. The positive bias targets what our ablations point to as the main cause of a vanishing subject, too little attention mass on its text tokens, rather than weight interference: an emphasis bias without any layout already recovers about half of the gain (\cref{sec:ablation}).

\subsection{Cyclic swap guidance}
\label{sec:swap}
The gating and the bias decide where each subject goes, but the two regions still read the same scene tokens, the shared part $W_0$ and the shared trigger, and they drift towards similar features (\cref{fig:mechanism}): a hybrid object, two copies of one subject, or one subject with a faded partner. We counteract this by guidance against a \emph{wrongly bound} prediction. Let $\pi(j)=j{+}1 \bmod m$ and let $\Omega^{\pi}$ bind subject $s_j$ to region $\Omega_{\pi(j)}$. With $v=v_\theta(x_i;y,\Omega)$ and $v^{\pi}$ the prediction with the bias of \cref{eq:bias} computed for $\Omega^{\pi}$ (gating unchanged),
\begin{equation}
\tilde v=v+w\,M\odot(v-v^{\pi}),\qquad M=\mathbb{1}\big[u\in\textstyle\bigcup_j\Omega_j\big],
\label{eq:swap}
\end{equation}
applied at steps $i\in\{0,2,4,6,8\}$ with $w{=}2$; background tokens are unchanged.

\paragraph{Why it counteracts homogenization.} \Cref{eq:swap} has the form of classifier-free guidance~\cite{cfg} with a degraded condition~\cite{cdg,autoguidance,pag}: the negative keeps the prompt, the adapter and the attention budget, and only binds each subject to a wrong region. In the idealized reading of guidance as sampling from $p(x\,|\,\Omega)\,[p(x\,|\,\Omega)/p(x\,|\,\Omega^{\pi})]^{w}$, the tilt rewards images by how much more likely they are under the correct binding than under the permuted one. An image whose regions show the same content, such as a hybrid or two copies, is nearly invariant under the permutation, so its log ratio is close to zero and it gains nothing; an image in which each region shows its own, distinct subject gains $w\log[p(x|\Omega)/p(x|\Omega^{\pi})]>0$. Locally, everything the two conditions share, namely the scene, the lighting, $W_0$ and the common trigger features, cancels in $v-v^{\pi}$, so the update acts only on features that discriminate a region's subject from its neighbor's. The cyclic shift is a derangement for every $m\ge2$, so one extra pass per guided step gives every region a negative, whatever the number of subjects; for $m{=}2$ it is a swap. Swapping only the bias, not the LoRA gating, keeps the negative pass close to the positive one, so the difference isolates the binding. The cost is five extra transformer calls ($28\to33$).

\begin{algorithm}[t]
\caption{Merge once, route early (Stage~2 sampling)}
\label{alg:main}
\small
\begin{algorithmic}[1]
\Require factored merge $\{W_0,W_1..W_K\}$ per layer (\cref{eq:split}); prompt $y$ with subjects $S$, token spans $\tau_j$; regions $\Omega_j$; $N_s{=}28$, $\gamma{=}0.3$, $p_0{=}2$, $w{=}2$, $\mathcal{G}{=}\{0,2,4,6,8\}$
\State $x_0\sim\mathcal{N}(0,I)$; $\Omega^{\pi}_{\pi(j)}\gets\Omega_j$ with $\pi(j)=j{+}1 \bmod m$
\For{$i=0,\dots,N_s{-}1$}
  \State gate $\gets$ (layout gating of \cref{sec:route}) \textbf{if} $i<\mathrm{round}(\gamma N_s)$ \textbf{else} all parts on all tokens
  \State $p_i\gets p_0\cos(\frac{\pi}{2}\frac{i}{N_s})$; build $\beta_i$ from $(\Omega,\tau)$ by \cref{eq:bias}
  \State $v\gets v_\theta(x_i;\,y,\ \text{gate},\ \beta_i)$
  \If{$i\in\mathcal{G}$}
    \State $v^{\pi}\gets v_\theta(x_i;\,y,\ \text{gate},\ \beta_i^{\pi})$ \Comment{bias from $(\Omega^{\pi},\tau)$}
    \State $v\gets v+w\,M\odot(v-v^{\pi})$ \Comment{\cref{eq:swap}}
  \EndIf
  \State $x_{i+1}\gets$ flow-matching Euler step$(x_i,v)$
\EndFor
\State \Return decode$(x_{N_s})$
\end{algorithmic}
\end{algorithm}

\section{Experiments}
\label{sec:exp}

\begin{figure*}[t]
\centering
\includegraphics[width=0.84\textwidth]{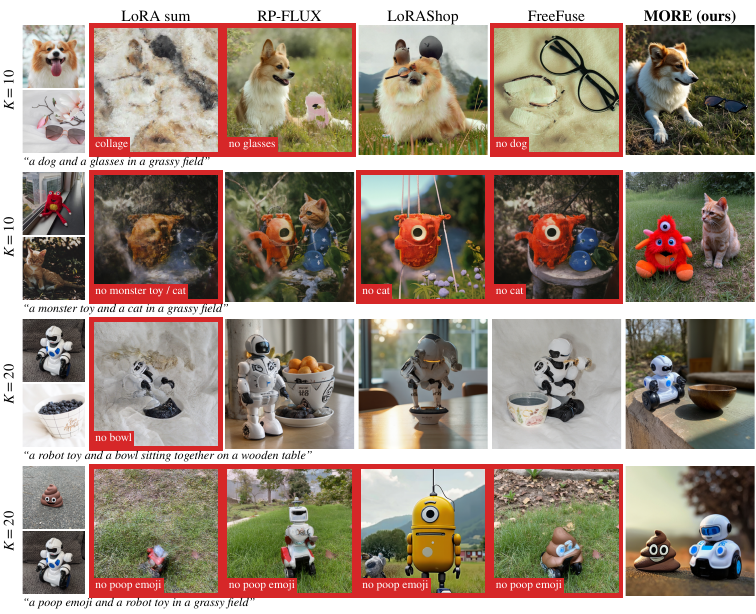}
\vspace{-0.9em}
\captionof{figure}{\textbf{Qualitative comparison} on held-out seeds ($K{=}10$ top, $K{=}20$ bottom; references at the left). Red frames mark failures, as found by the metric of \cref{sec:setup}: a missing subject or a collage. The baselines load only the prompt's two LoRAs; ours carries all $K$. The dominant failure is a missing subject, \eg, the poop emoji in row~4 for every baseline; LoRA sum also escapes into collages.}
\label{fig:qual}
\vspace{0.4em}
\footnotesize
\setlength{\tabcolsep}{3.1pt}
\captionof{table}{\textbf{Comparison on the held-out pair benchmark} (20 prompts $\times$ seeds 6--11, $n{=}120$ per cell). \col: collage rate; \bothC/\minC/\meanC: coexistence and per-subject identity with collages counted as failures (\cref{sec:setup}); HPS: HPSv2.1; IR: ImageReward; CLIP-T/I: text and reference similarity; time: seconds per image at $K{=}10$. ``LoRAs'': adapters loaded at inference; all baselines receive the prompt's own two LoRAs (oracle selection), ours carries the whole pool in one rank-$r$ adapter. ``Layout'': oracle left/right layout, automatic masks, or none. Paired differences of \bothC{} to ours are in \cref{sec:main}.}
\label{tab:main}
\vspace{-0.3em}
\resizebox{\textwidth}{!}{%
\begin{tabular}{@{}lcc|cccccccc|ccccc|c@{}}
\toprule
& & & \multicolumn{8}{c|}{$K=10$} & \multicolumn{5}{c|}{$K=20$} & \\
Method & LoRAs & Layout & \col$\downarrow$ & \bothC$\uparrow$ & \minC$\uparrow$ & \meanC$\uparrow$ & HPS$\uparrow$ & IR$\uparrow$ & CLIP-T$\uparrow$ & CLIP-I$\uparrow$ & \bothC$\uparrow$ & \minC$\uparrow$ & \meanC$\uparrow$ & HPS$\uparrow$ & IR$\uparrow$ & time$\downarrow$ \\
\midrule
\textcolor{black!55}{FLUX.1-dev (no LoRA)} & \textcolor{black!55}{--} & \textcolor{black!55}{--} & \textcolor{black!55}{.083} & \textcolor{black!55}{.267} & \textcolor{black!55}{.083} & \textcolor{black!55}{.199} & \textcolor{black!55}{.256} & \textcolor{black!55}{$-$0.30} & \textcolor{black!55}{.235} & \textcolor{black!55}{.607} & \textcolor{black!55}{.233} & \textcolor{black!55}{.066} & \textcolor{black!55}{.207} & \textcolor{black!55}{.263} & \textcolor{black!55}{$-$0.49} & \textcolor{black!55}{--} \\
LoRA sum & 2 & -- & .200 & .017 & .028 & .087 & .117 & $-$2.03 & .165 & .582 & .033 & .028 & .074 & .128 & $-$2.04 & 4.3 \\
LoRA Switch~\cite{multilora} & 2 & -- & .083 & .125 & .069 & .179 & .172 & $-$1.42 & .199 & .624 & .158 & .081 & .185 & .179 & $-$1.33 & 4.2 \\
LoRA Composite~\cite{multilora} & 2 & -- & .092 & .108 & .077 & .175 & .169 & $-$1.57 & .196 & .621 & .183 & .089 & .195 & .186 & $-$1.26 & 8.4 \\
SSR-Merge~\cite{ssr} & all $K$ & -- & .058 & .083 & .059 & .183 & .193 & $-$1.44 & .211 & .618 & .017 & .037 & .139 & .175 & $-$1.93 & 4.4 \\
RP-FLUX~\cite{rpflux} & 2 & oracle & .133 & .433 & .140 & .253 & .240 & \underline{0.07} & .251 & .664 & .308 & .110 & .266 & .238 & $-$0.42 & 7.7 \\
LoRAShop~\cite{lorashop} & 2 & auto & .083 & .250 & .091 & .220 & .223 & $-$0.68 & .233 & .624 & .242 & .080 & .220 & .225 & $-$0.78 & 30.7 \\
FreeFuse~\cite{freefuse} & 2 & auto & .058 & .508 & .158 & .302 & .212 & $-$0.60 & .238 & \underline{.674} & .358 & .141 & .323 & .217 & $-$0.68 & 9.0 \\
\midrule
MORE, static merge & all $K$ & -- & .025 & .225 & .089 & .273 & .248 & $-$0.58 & .226 & .654 & .150 & .077 & .321 & .241 & $-$0.88 & 4.4 \\
MORE, auto layout & all $K$ & auto & .000 & \underline{.700} & \underline{.167} & \underline{.318} & \underline{.285} & 0.75 & \underline{.266} & .672 & \underline{.550} & \underline{.181} & \underline{.367} & \underline{.288} & 0.65 & -- \\
\textbf{MORE (ours)} & all $K$ & oracle & \textbf{.008} & \textbf{.942} & \textbf{.207} & \textbf{.339} & \textbf{.301} & \textbf{1.32} & \textbf{.279} & \textbf{.681} & \textbf{.783} & \textbf{.245} & \textbf{.379} & \textbf{.307} & \textbf{1.51} & 6.9 \\
\bottomrule
\end{tabular}}
\vspace{-0.8em}
\end{figure*}

\subsection{Setup}
\label{sec:setup}
\paragraph{Pool and benchmark.} We train one rank-32 LoRA per subject for 20 DreamBooth~\cite{dreambooth} subjects (backpack, stuffed animal, cat, dog, toy, monster toy, glasses, robot toy, sneaker, vase, rc car, teapot, wolf plushie, clock, candle, bowl, can, boot, sloth plushie, poop emoji) on all 342 linear layers of FLUX.1-dev, each with prompt ``a \sks{} $\langle$class$\rangle$''. The $K{=}10$ pool holds the first ten, the $K{=}20$ pool all. Per pool, the pair benchmark has 10 subject pairs $\times$ 2 templates (``a \sks{} $A$ and a \sks{} $B$ sitting together on a wooden table'' / ``\dots{} in a grassy field''). We sample at $512{\times}512$, 28 steps, guidance 3.5, seed $1000\,\sigma{+}i$ for prompt $i$ and seed index $\sigma$. \emph{All design choices and hyperparameters were selected on seeds 0--5; all reported numbers use the held-out seeds 6--11} ($n{=}120$ per method and pool) unless marked ``dev''. The default layout puts the first-mentioned subject in the left half and the second in the right half. Crosstalk (\cref{sec:intro}) is the output energy of the $K{-}1$ foreign LoRAs relative to the own LoRA on each subject's non-padding prompt tokens, from the calibration inputs; over subjects and layers it ranges from $3.9\times$ to $16.1\times$ (median $7.9\times$) at $K{=}10$ and from $7.9\times$ to $30.4\times$ (median $17.0\times$) at $K{=}20$.

\paragraph{Metrics.} Each subject is localized with OWLv2~\cite{owlv2} (query: its class; the $m$ boxes are the disjoint assignment maximizing the summed scores, so one hybrid cannot count for two subjects), and its square crop is compared with the subject's reference images by DINOv2~\cite{dinov2} cosine. A CLIP~\cite{clip} zero-shot split-screen test flags collages (\col). \bothC{} is the fraction of images in which every subject is detected (score ${\ge}0.2$) and the image is not a collage, the complement of a subject-collapse rate~\cite{collapse}; \minC{} and \meanC{} are the minimum and mean crop identity over the subjects, with collages scored $0$. \bothC{} measures coexistence, \meanC{} identity. We also report HPSv2.1~\cite{hpsv2}, ImageReward~\cite{imagereward}, CLIP-T (text alignment, prompt without \sks) and CLIP-I (reference similarity). Uncertainties are paired bootstrap standard errors over the same prompts and seeds.

\paragraph{Baselines.} All baselines receive only the prompt's two LoRAs, which is easier than our setting, where all $K$ are loaded. LoRA sum adds both at weight 1. LoRA Switch and Composite~\cite{multilora} are re-implemented on FLUX (switching every 5 steps; mean of per-LoRA predictions). RP-FLUX~\cite{rpflux} runs the official code with our oracle layout, regional prompts ``a \sks{} $\langle$class$\rangle$'' plus the scene, $-\infty$ masks for 10 steps and base ratio 0.3; we report LoRA weight 0.5, since weight 1 collapses (\bothC{} .050 / .042, \col{} .41). LoRAShop~\cite{lorashop} and FreeFuse~\cite{freefuse} run their official code and infer subject masks from the base model. SSR-Merge~\cite{ssr} is our reproduction of the published merge (one prompt per subject, $\lambda{=}10^{-4}$) over the whole pool. For a comparison without a layout prior we also run \emph{MORE, auto layout}: subject maps from a 5-step base-model pass (double block 18) are thresholded into two regions, as FreeFuse and LoRAShop do. Times are wall-clock per image on one GPU from our generation logs, including the extra passes.

\begin{figure}[t]
\centering
\includegraphics[width=\linewidth]{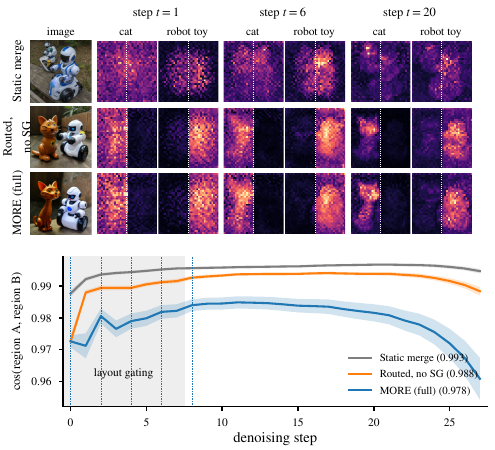}
\vspace{-1.8em}
\caption{\textbf{Mechanism} ($K{=}10$, held-out seed). Top: attention mass of the image tokens on each subject's text tokens (double block 18, bias included) at steps 1, 6, 20 for the static merge, routing without swap guidance, and ours; the dotted line is the layout boundary. Bottom: cosine similarity of the mean block-18 features of the two regions over the steps (mean $\pm$ s.e.\ over 20 images; legend: mean over steps 0--7). Gating and bias separate the maps; swap guidance (its steps dotted) additionally pulls the region features apart.}
\label{fig:mechanism}
\vspace{-1.0em}
\end{figure}

\subsection{Comparison with the state of the art}
\label{sec:main}
\Cref{tab:main} and \cref{fig:qual} summarize the comparison. At $K{=}10$ our pipeline keeps both subjects in $94.2\%$ of the images against $50.8\%$ for FreeFuse, the strongest baseline ($\Delta$\bothC{} $=0.433\pm0.050$), $43.3\%$ for RP-FLUX with the same oracle layout ($0.508\pm0.044$) and $25.0\%$ for LoRAShop ($0.692\pm0.043$). At $K{=}20$ the gap to FreeFuse is $0.425\pm0.053$ ($78.3\%$ vs.\ $35.8\%$). Ours is also best on identity (\minC, \meanC), human-preference and text metrics, and our collage rate is the lowest ($0.8\%$). The large ImageReward gap ($1.32$ vs.\ $-0.30$ for the base model) mostly reflects composition, since all arms are scored with the same prompt: base-model images score $+0.78$ when both subjects are detected and $-0.71$ when one is missing (87 of 120), and the base model with only our attention bias already reaches $1.30$. The LoRA-switching schemes and the static merges compose no better than the base model, which shows that coexistence is not a weight-space property. \emph{MORE, static merge} is still the best static adapter on identity (\meanC{} $.273$ vs.\ $.183$ for SSR-Merge), the prerequisite for the routed parts to carry identity.

\paragraph{Without a layout prior.} With automatic layouts our \bothC{} drops by about $0.24$ at both $K$ ($0.700$ / $0.550$) but stays above FreeFuse and LoRAShop, which use the same kind of base-model masks, with better identity and ImageReward. Mask inference is thus a bottleneck shared by all automatic methods, not a requirement of our pipeline.

\paragraph{Cost.} One rank-$8$/$16$ adapter replaces $K$ rank-32 LoRAs; requests differ only in routing. Sampling takes $6.9$\,s per image at both $K$ ($1.55\times$ the static merge, from the five swap-guidance passes), against $9.0$\,s / $10.5$\,s for FreeFuse and $30.7$\,s / $34.2$\,s for LoRAShop.

\begin{table}[t]
\centering
\footnotesize
\setlength{\tabcolsep}{2.6pt}
\caption{\textbf{Ablations} (held-out, $n{=}120$; \bothC{} / \minC{} / \meanC). (a) Merge, each routed by the full Stage~2. (b) Routing, all on our merge unless noted; ``bias'' = \cref{eq:bias} without the image--image term; ``+ii+sg'' adds the image--image term and swap guidance; ``gating'' = layout gating in all steps.}
\label{tab:ablation}
\vspace{-0.6em}
\resizebox{\columnwidth}{!}{%
\begin{tabular}{@{}lccc|ccc@{}}
\toprule
& \multicolumn{3}{c|}{$K=10$} & \multicolumn{3}{c}{$K=20$} \\
& \bothC & \minC & \meanC & \bothC & \minC & \meanC \\
\midrule
\multicolumn{7}{@{}l}{\textit{(a) merge under full routing}} \\
raw own two LoRAs & .592 & .184 & .271 & .367 & .129 & .232 \\
task arithmetic (avg) & .900 & .197 & .292 & .750 & .198 & .301 \\
SSR-Merge, published calib. & .525 & .134 & .230 & .342 & .105 & .208 \\
SSR-Merge, our calib. & .883 & \textbf{.211} & .343 & .750 & .220 & .340 \\
MORE w/o de-crowding & \textbf{.967} & \textbf{.215} & \textbf{.347} & \textbf{.808} & \textbf{.250} & .373 \\
MORE & .942 & .207 & .339 & .783 & .245 & \textbf{.379} \\
\midrule
\multicolumn{7}{@{}l}{\textit{(b) routing}} \\
static merge & .225 & .089 & .273 & .150 & .077 & .321 \\
static merge + bias & .742 & .159 & .335 & .542 & .207 & .368 \\
\ \ + ii + sg & .825 & .202 & \textbf{.362} & .725 & .240 & .376 \\
gating + bias & .817 & .175 & .296 & .592 & .217 & .348 \\
base FLUX + bias & .783 & .177 & .268 & .758 & .168 & .272 \\
\ \ + ii + sg & .875 & .198 & .275 & .725 & .172 & .264 \\
\textbf{full} (early gating + bias + ii + sg) & \textbf{.942} & \textbf{.207} & .339 & \textbf{.783} & \textbf{.245} & \textbf{.379} \\
\bottomrule
\end{tabular}}
\vspace{-1.0em}
\end{table}

\subsection{Ablations}
\label{sec:ablation}
\paragraph{Merge (\cref{tab:ablation}a).} Under the same routing, the raw two LoRAs are far worse than any calibrated merge ($.592$ vs.\ $.942$ \bothC{} at $K{=}10$, with $15\%$ collages): a token-routed sum of two LoRAs trained on the same trigger still overshoots, while the merge is fitted to each subject's own response. The calibration is the largest merge-side factor: the published SSR-Merge statistics give $.525$ / $.342$, the same router on our statistics $.883$ / $.750$. Relative ridge and fixed rank matter mainly at $K{=}20$ (\meanC{} $+.039\pm.011$ over SSR-Merge with our calibration). Task arithmetic composes almost as well but loses identity ($\Delta$\meanC{} $-.047\pm.011$ / $-.079\pm.010$). Shared-direction damping~\cite{pico} is \emph{not} significant ($+.025\pm.019$ / $+.025\pm.028$ \bothC{} without it); we keep it only for consistency with the static merge. A crosstalk-weighted least-squares variant raised HPS by ${\approx}0.01$ but lowered \minC/\meanC{} by $0.02$--$0.04$ at $K{=}20$, so we dropped it.

\paragraph{Routing (\cref{tab:ablation}b).} The bias alone lifts the static merge from $.225$ to $.742$ \bothC{} at $K{=}10$; on the base model it does too ($.783$), but there identity stays at the base level (\meanC{} $.268$), and at $K{=}20$ it composes almost as well as ours ($.758$) but with much lower identity (\minC{} $.168$ vs.\ $.245$); the merged adapter is what carries identity (ours vs.\ base FLUX with the same routing: $\Delta$\meanC{} $+.064\pm.013$ at $K{=}10$); on the base model the corrections even produce collages at $K{=}20$ ($12.5\%$). Gating, the image--image term and swap guidance each add composition: full vs.\ gating + bias gives $\Delta$\bothC{} $+.125\pm.039$ / $+.192\pm.040$ and $\Delta$\meanC{} $+.043\pm.011$ / $+.031\pm.009$ at $K{=}10$ / $20$. The ungated static merge with the same corrections has the best \meanC{} at $K{=}10$ ($.362$, ours $-.023\pm.011$) but $11.7$ points lower \bothC. On dev seeds, adding parts to gating + bias ($.858$/$.185$/$.297$): early-only gating $.858$/$.186$/$.317$, $+$ii $.850$/$.175$/$.319$, $+$swap guidance $w{=}1$ $.917$/$.204$/$.339$, $w{=}2$ $.942$/$.218$/$.349$; $w{=}3$ and $\gamma{=}0.2$ are no better.

\begin{figure*}[t]
\centering
\includegraphics[width=\textwidth]{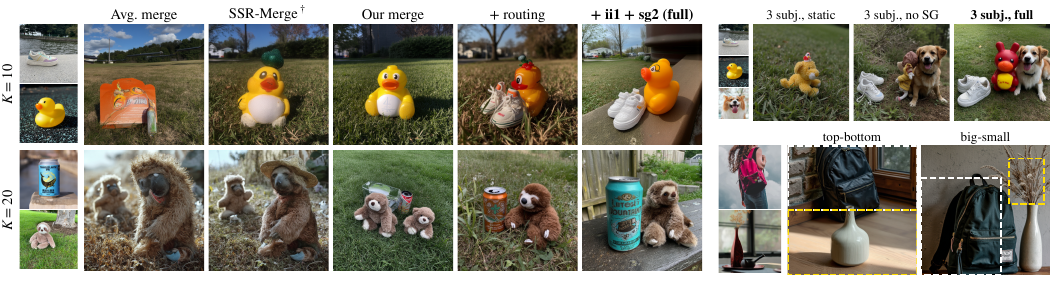}
\vspace{-1.8em}
\caption{\textbf{Ablation and generalization} (held-out seeds). Left: from static merges (task arithmetic, SSR-Merge$^\dagger$ with our calibration, our merge) to routing (early gating + bias) and the full pipeline, at $K{=}10$ and $K{=}20$. Right top: three subjects, static merge vs.\ routing without and with swap guidance. Right bottom: the same prompt with top-bottom and big-small layouts (dashed boxes).}
\label{fig:ablation}
\vspace{-1.0em}
\end{figure*}

\paragraph{Attention design (dev).} Hard masks hurt on our merge: RP-FLUX-style $-\infty$ blocking of cross-subject entries for the first $30\%$ of the steps lowers raw both-detected from $.275$ (static) to $.133$, and a finite block of $-4$ to $.075$, mostly through collages; the positive own bias without any blocking reaches $.692$. Without a layout, the same $+2$ on all image tokens reaches $.500$; $p_0{=}1$ halves the gain and $p_0{=}4$ destroys the images. The bias must act in all blocks: restricting it to the middle double blocks (4--12) removes the gain, and the double or the single blocks alone each give about half of it. Stronger image--image repulsion ($-1.5p_i$) causes $33\%$ collages; a region-swapped NAG~\cite{nag} negative adds nothing on our merge.

\paragraph{Mechanism (\cref{fig:mechanism}).} With the static merge both subjects' attention spreads over the whole image and the regions' features are almost identical (cosine $.993$ over steps 0--7). Routing separates the attention maps ($.988$), and swap guidance lowers the feature similarity further, in the gated steps ($.978$) and throughout sampling. This separation is in feature space: our detected crops look \emph{more} alike than the baselines' (DINOv2 cosine $.72$ vs.\ $.55$ for FreeFuse; confounded by missing subjects), and swap guidance does not lower it, so we claim no crop-level effect (\cref{sec:app_conf}).

\begin{table}[t]
\centering
\footnotesize
\setlength{\tabcolsep}{2.8pt}
\caption{\textbf{Generalization} ($K{=}10$, held-out, $n{=}120$). (a) Layouts: left-right, top-bottom, big-small (a large and a small box); hit: all subjects detected inside their regions. (b) Three subjects, \allC{} = every subject present, no collage. (c) Blind A/B study (one rater, 50 trials, 10 per baseline; ties excluded from the win rate, 95\% Wilson interval).}
\label{tab:general}
\vspace{-0.6em}
\resizebox{\columnwidth}{!}{%
\begin{tabular}{@{}lcccccc@{}}
\toprule
\multicolumn{7}{@{}l}{\textit{(a) layouts}} \\
& \bothC & \minC & \meanC & hit & HPS & IR \\
\midrule
left-right & .942 & .207 & .339 & .933 & .301 & 1.32 \\
top-bottom & .733 & .192 & .299 & .642 & .281 & 1.07 \\
big-small & .475 & .124 & .271 & .117 & .263 & 0.21 \\
\midrule
\multicolumn{7}{@{}l}{\textit{(b) three subjects}} \\
& \allC & \minC & \meanC & \col & HPS & IR \\
\midrule
raw own LoRAs, static & .000 & $-$.005 & .020 & .383 & .072 & $-$2.27 \\
SSR-Merge, static & .017 & .027 & .164 & .042 & .193 & $-$1.64 \\
our merge, static & .100 & .045 & .228 & .042 & .260 & $-$0.39 \\
MORE w/o swap guidance & .608 & .095 & .258 & .058 & .292 & 1.15 \\
\textbf{MORE} & \textbf{.692} & \textbf{.115} & \textbf{.278} & \textbf{.033} & .293 & \textbf{1.35} \\
\midrule
\multicolumn{7}{@{}l}{\textit{(c) user study vs.\ MORE (overall question)}} \\
& \multicolumn{2}{c}{win} & tie & loss & \multicolumn{2}{c}{win w/o ties} \\
\midrule
all 5 baselines ($n{=}49$) & \multicolumn{2}{c}{.959} & .041 & .000 & \multicolumn{2}{c}{1.00 [.92, 1]} \\
RP-FLUX ($n{=}10$) & \multicolumn{2}{c}{.800} & .200 & .000 & \multicolumn{2}{c}{1.00 [.68, 1]} \\
\bottomrule
\end{tabular}}
\vspace{-1.0em}
\end{table}

\subsection{Layouts, three subjects and user study}
\label{sec:general}
\paragraph{Layouts.} Only the region masks change (\cref{tab:general}a, \cref{fig:ablation}). Top-bottom keeps both subjects in $73.3\%$ of the images with ImageReward $1.07$. Big-small (a large left box and a small top-right box) is the weak spot: \bothC{} $.475$, and both subjects land in their boxes in only $11.7\%$ of the images; the large subject tends to spill over the small region (\cref{sec:conclusion}).

\paragraph{Three subjects.} With the cyclic shift, the pipeline applies unchanged to $m{=}3$ (three vertical strips, 20 triples). All three subjects coexist in $69.2\%$ of the images against $10.0\%$ for our static merge ($\Delta$\allC{} $+.592\pm.045$) and $0\%$ for the raw LoRAs; swap guidance adds $8.4$ points \allC{} and $0.20\pm0.08$ ImageReward, and all subjects land in their strips in $65\%$ of the images.

\paragraph{User study.} One rater (the author) answered 50 blind A/B trials of random held-out $K{=}10$ images, ten per baseline (FreeFuse, LoRAShop, RP-FLUX, SSR-Merge, LoRA sum), with randomized sides and three questions: both identities kept, text match, overall. Ours was never judged worse (\cref{tab:general}c); all ties were against RP-FLUX, and the identity and text questions give the same picture ($n{=}47$, win $.957$, tie $.043$). Seven of 150 answers were left blank and are excluded. A single, non-naive rater is no substitute for a crowd study; all 50 trials and answers are in \cref{sec:app_us}.

\section{Limitations and conclusion}
\label{sec:conclusion}
\paragraph{Limitations.} The pipeline needs a layout; with automatic masks \bothC{} drops by $0.24$. Small regions are its weak spot: in the big-small layout both subjects land in their boxes in only $11.7\%$ of the images (\bothC{} $.475$), presumably because a small region collects too little attention mass. The bias also raises color saturation (by $0.08$ on average at $p_0{=}2$, with unchanged luminance), routing adds $2.4$\,s per image over the static merge, and prompts are shared between our dev and test seeds; and identity rather than coexistence remains the bottleneck at $K{=}20$ (\minC{} $.245$). Swap guidance separates region features, not crop-level identity similarity. All baselines receive the prompt's own LoRAs, and LoRAtorio~\cite{loratorio} is missing; our pool shares one trigger word, and the user study has a single rater who knows the method (though not the sides), so a crowd study (${\ge}30$ raters) is needed before preference claims.

\paragraph{Conclusion.} Composing subjects from a large LoRA pool fails on FLUX for two separable reasons, crosstalk among the pool's deltas and coexistence collapse of a token-agnostic adapter. MORE addresses both without training, merging once and routing only early; it keeps $94\%$/$78\%$ of the pairs complete at $K{=}10$/$20$ and extends unchanged to three subjects.

{
    \small
    \bibliographystyle{ieeenat_fullname}
    \bibliography{main}
}

\clearpage
\appendix
\section{Protocol details}
\label{sec:app_protocol}
\paragraph{Arms.} Our full configuration is the routed merge with layout gating for $\mathrm{round}(0.3\cdot28){=}8$ steps, the bias of \cref{eq:bias} with $p_0{=}2$ and cosine decay in all 57 blocks and all steps, image--image repulsion $-p_i/2$, and swap guidance ($w{=}2$, steps $\{0,2,4,6,8\}$, bias-only swap). Stage~1 uses relative ridge $\rho{=}0.05$/$0.1$, rank $r{=}8$/$16$, energy gain to the own-LoRA energy, and the calibration of \cref{sec:merge}. Each factored adapter is checked against its static weights (relative error ${\le}6{\times}10^{-7}$).
\paragraph{Selection.} All choices were made on seeds 0--5 (\cref{sec:app_hparam}): the Stage-2 choices (gating length, $p_0$, decay, blocking, repulsion, $w$, swap target) on the same 20 pair prompts, the Stage-1 choices (rank, ridge, gain, calibration mixing, de-crowding) on a separate single-subject benchmark; the guided steps $\{0,2,4,6,8\}$ were fixed and not tuned. Seeds 6--11 were generated once for the final configuration and every ablation arm, and all tables of the main text report them. The pair prompts are therefore shared between dev and test; only the noise differs.
\paragraph{Metrics.} OWLv2 queries are the class words; the $m$ boxes are chosen jointly to be disjoint (IoU $<0.5$) with maximal score sum. Collages are flagged when CLIP ViT-B/32 assigns probability ${>}0.5$ to split-screen prompts (\eg, ``a collage of two separate pictures with a vertical dividing line'') against single-photo prompts; top-bottom layouts use horizontal-split prompts and triples the maximum of vertical and three-panel prompts. HPSv2.1, ImageReward v1.0, CLIP-T and CLIP-I (ViT-L/14) use the prompt with \sks{} removed. ImageReward is reported in its native units (roughly standardized, unbounded). All arms use identical cached text embeddings, 28 steps, guidance 3.5 and $512{\times}512$, and are scored on the raw images with the same prompt; collages are not penalized in HPS/IR. Split by whether both subjects are detected, ImageReward at $K{=}10$ is $+0.78$ / $-0.71$ for the base model, $+0.17$ / $-1.43$ for FreeFuse, $+0.98$ / $-0.75$ for RP-FLUX, $+1.57$ / $+0.07$ for the base model with our bias and $+1.43$ / $-0.89$ for ours: a missing subject costs about $1.5$ points for every method, and the base model's images are of high quality but usually show one subject only.

\section{Derivations}
\label{sec:app_deriv}
This section spells out the short derivations behind \cref{sec:merge,sec:route,sec:swap}. They are standard linear algebra and an idealized reading of guidance, given to make the method reproducible and its approximations explicit; they are not new theory. Throughout, $z=\Acomb x\in\mathbb{R}^{Kr_0}$, $E_k$ is the diagonal $0/1$ projector onto block $k$ of $z$ (so $B_kA_kx=\Bcomb E_kz$), and per subject $G_k=\sum_{x\in\mathcal{X}_k}zz^{\top}$, so that $G=\sum_kG_k$ and $Q=\sum_kE_kG_k$ in \cref{eq:router}.

\subsection{Router}
\label{sec:app_router}
\paragraph{Normal equations.} With $\Delta W=\Bcomb R\Acomb$ and a ridge on the merged output map $M=\Bcomb R$,
\begin{equation}
J(R)=\sum_k\sum_{x\in\mathcal{X}_k}\big\|\Bcomb(R-E_k)z\big\|^2+n\lambda\,\|\Bcomb R\|_F^2 .
\label{eq:app_obj}
\end{equation}
Using $\sum_{x\in\mathcal{X}_k}\|\Bcomb(R-E_k)z\|^2=\mathrm{tr}\big(\Bcomb(R-E_k)G_k(R-E_k)^{\top}\Bcomb^{\top}\big)$, the gradient is
\begin{equation}
\tfrac12\nabla_RJ=\Bcomb^{\top}\Bcomb\big(RG-Q+n\lambda R\big).
\end{equation}
Hence every $R$ with $R(G+n\lambda I)=Q$ is a minimizer, \ie, $R=Q(G+n\lambda I)^{-1}=\tfrac{Q}{n}(\tfrac{G}{n}+\lambda I)^{-1}$, which is \cref{eq:router}; it is the unique minimizer when $\Bcomb$ has full column rank (generically the case, as $Kr_0\le640$ is far below $d_{\mathrm{out}}$). For $\lambda\to0$ and invertible $G$, $R=QG^{-1}$, the router of SSR-Merge~\cite{ssr}. Since the unrestricted RegMean objective evaluated at $\Delta W=\Bcomb R\Acomb$ is the first term of \cref{eq:app_obj}, this $R$ is RegMean~\cite{regmean} restricted to the LoRA subspace.

\paragraph{The router corrects crosstalk.} Task arithmetic is $R=I$. Writing $G=\sum_kE_kG_k+\sum_k(I-E_k)G_k$,
\begin{equation}
R-I=-\Big(\underbrace{\textstyle\sum_k(I-E_k)G_k}_{C_{\times}}+n\lambda I\Big)\big(G+n\lambda I\big)^{-1}.
\label{eq:app_cross}
\end{equation}
$C_{\times}$ collects the features that subject $k$'s inputs produce in the \emph{other} LoRAs' A-blocks; it is the crosstalk of the pool. If every subject's inputs excited only its own block ($C_{\times}{=}0$), the router would reduce to $R=G(G+n\lambda I)^{-1}\to I$; for $K{=}1$, $C_{\times}{=}0$ always and the single LoRA is recovered. The deviation from task arithmetic is therefore driven by the crosstalk on the calibration inputs, which is why calibration is the largest merge-side factor (\cref{tab:ablation}).

\paragraph{Relative ridge.} Let $G/n=V\,\mathrm{diag}(\mu_i)V^{\top}$. Then $R=\tfrac{Q}{n}V\,\mathrm{diag}\big(1/(\mu_i+\lambda)\big)V^{\top}$, \ie, the ridge damps the inverse in feature directions with $\mu_i\lesssim\lambda$. With $\lambda=\rho\,\mathrm{tr}(G/n)/(Kr_0)=\rho\,\bar\mu$, a direction is damped when its variance falls below a fraction $\rho$ of the mean feature variance. If a layer's features are scaled, $z\to cz$, then $G\to c^2G$, $Q\to c^2Q$, $\lambda\to c^2\lambda$, and $R$ is unchanged; the result is also independent of the number of calibration tokens $n$. An absolute $\lambda$ instead acts as a strong prior towards $R{=}0$ on low-scale layers and as no regularization on high-scale ones.

\subsection{Factored rank-\texorpdfstring{$r$}{r} truncation}
\label{sec:app_svd}
Let $M=\Bcomb R=U_mS_mV_m^{\top}$ and $\Acomb=U_aS_aV_a^{\top}$ be thin SVDs ($U_m\in\mathbb{R}^{d_{\mathrm{out}}\times Kr_0}$, $V_a\in\mathbb{R}^{d_{\mathrm{in}}\times Kr_0}$, both with orthonormal columns), and let $C=S_mV_m^{\top}U_aS_a=U_cS_cV_c^{\top}$, a $Kr_0{\times}Kr_0$ matrix. Then
\begin{equation}
M\Acomb=U_m\,C\,V_a^{\top}=(U_mU_c)\,S_c\,(V_aV_c)^{\top},
\end{equation}
and since $U_mU_c$ and $V_aV_c$ have orthonormal columns, this is an SVD of $M\Acomb$. By Eckart--Young, its best rank-$r$ approximation in Frobenius norm keeps the first $r$ terms, with error $\|M\Acomb-\Delta W_r\|_F^2=\sum_{i>r}\sigma_{c,i}^2$. Splitting $S_{c,r}=s\,s$ with $s=S_{c,r}^{1/2}$ gives the factors
\begin{equation}
\begin{aligned}
B_r&=U_mU_{c,r}\,s,\qquad A_r=s\,V_{c,r}^{\top}V_a^{\top}=T\Acomb,\\
T&=s\,V_{c,r}^{\top}S_a^{-1}U_a^{\top},
\end{aligned}
\end{equation}
where $T\Acomb=sV_{c,r}^{\top}S_a^{-1}U_a^{\top}U_aS_aV_a^{\top}=A_r$ ($S_a^{-1}$ is a pseudo-inverse if $\Acomb$ is rank-deficient; the identity still holds because the right singular vectors of $C$ with nonzero singular value have no component on the zero columns of $S_a$). The cost is that of SVDs of $d{\times}Kr_0$ and $Kr_0{\times}Kr_0$ matrices; no $d_{\mathrm{out}}{\times}d_{\mathrm{in}}$ matrix is formed. The truncation is in weight space (Frobenius norm of $\Delta W$). Because $A_r=T\Acomb$, the compressed merge still reads the LoRA-A features $z$, which is what makes the split of \cref{eq:split} possible.

\paragraph{Energy gain.} For $\Delta W_r=M_rz$ with $M_r=B_rT$, the calibration energies are quadratic forms in the Gram matrices,
\begin{equation}
E_r=\sum_k\mathrm{tr}\big(M_rG_kM_r^{\top}\big),\qquad E_{\mathrm{own}}=\sum_k\mathrm{tr}\big(B_k\,G_k^{[kk]}B_k^{\top}\big),
\end{equation}
with $G_k^{[kk]}$ the $(k,k)$ block of $G_k$ and $B_k$ the original (undamped) LoRA factors, so $g=(E_{\mathrm{own}}/E_r)^{1/2}$ is computed without another pass over data. It undoes the norm loss of the ridge (the factor $\mu_i/(\mu_i{+}\lambda)$ above) and of the truncation, so that the merged adapter acts on the calibration inputs with the strength of the own LoRAs; one scalar per layer does not change the direction of $\Delta W_r$.

\paragraph{Shared-direction damping.} With $u_1,\dots,u_8$ the top left singular vectors of $\Bcomb$ and $\mathrm{share}_j=\frac1K\sum_k\|u_j^{\top}B_k\|/\|B_k\|_F$, directions with $\mathrm{share}_j>0.5$ are damped before \cref{eq:router}, $\Bcomb\leftarrow\Bcomb-\sum_j0.5\,\mathrm{share}_j\,u_ju_j^{\top}\Bcomb$, and $U_s=[u_j]_{\mathrm{share}_j>0.5}$ defines $P=U_sU_s^{\top}$ in \cref{eq:split}.

\subsection{Exact factorization and gating}
\label{sec:app_split}
Writing $T=[T_1,\dots,T_K]$ in column blocks of width $r_0$, block multiplication gives $T\Acomb x=\sum_kT_kA_kx$, and $P+(I{-}P)=I$ gives
\begin{equation}
\begin{aligned}
gB_rT\Acomb x&=gPB_rT\Acomb x+\textstyle\sum_kg(I{-}P)B_rT_kA_kx\\
&=W_0x+\textstyle\sum_kW_kx,
\end{aligned}
\end{equation}
which is \cref{eq:split}; no approximation is involved (numerically, relative error ${\le}6{\times}10^{-7}$, \cref{sec:app_protocol}). Stage~2 evaluates $\omega_0(u)W_0x_u+\sum_k\omega_k(u)W_kx_u$ per token $u$ with weights $\omega\in\{0,1\}$; all weights one is the static merge. In region $\Omega_j$ during gating, $\omega_{s_l}{=}0$ for the partners $l\neq j$ and all other weights are one, so the change with respect to the static merge is exactly
\begin{equation}
-\textstyle\sum_{l\neq j}W_{s_l}x_u=-g(I{-}P)B_r\sum_{l\neq j}T_{s_l}A_{s_l}x_u,
\end{equation}
the merged adapter's response to the partners' LoRA-A features of that token, which is the crosstalk a token of $\Omega_j$ receives from the other prompt subjects. Since all parts share $B_r$, the routed output is $g\,[PB_rT\Acomb x_u+(I{-}P)B_r\sum_k\omega_kT_kA_kx_u]$, so gating costs $K$ small $r{\times}r_0$ products per token on top of the static merge.

\subsection{Attention mass under the soft bias}
\label{sec:app_bias}
Fix a query $a\in\Omega_j$ in one attention head and let, before the bias, $\alpha$ be its softmax mass on the subject's text $\tau_j$, $\alpha_{\times}$ its mass on the other regions $\Omega_l$, $l\neq j$, and $1-\alpha-\alpha_{\times}$ the rest. Adding \cref{eq:bias} multiplies the unnormalized weights by $e^{p_i}$, $e^{-p_i/2}$ and $1$, so
\begin{equation}
\begin{aligned}
\alpha'&=\frac{e^{p_i}\alpha}{e^{p_i}\alpha+e^{-p_i/2}\alpha_{\times}+1-\alpha-\alpha_{\times}},\\
\frac{\alpha'}{1-\alpha'}&\ \ge\ e^{p_i}\,\frac{\alpha}{1-\alpha},
\end{aligned}
\label{eq:app_mass}
\end{equation}
with equality when $\alpha_{\times}{=}0$: the bias multiplies the odds of attending to the own subject's tokens by at least $e^{p_i}$, whatever the head's logits, and the image--image term increases the factor further. With $p_0{=}2$ and the cosine decay the factor is $7.4$ at step $0$, $6.1$ at step $8$ (end of gating), $4.1$ at step $14$ and $1.1$ at step $27$. Every key keeps a nonzero weight, so a region can still attend to the partner's tokens; a $-\infty$ mask instead sets that mass to zero. The same factor applies in the reverse direction to text queries in $\tau_j$ attending to $\Omega_j$. This is a per-layer statement about the logits that the model sees; its effect on the output is measured, not derived (\cref{sec:ablation,fig:mechanism}).

\subsection{Swap guidance as a tilt}
\label{sec:app_swap}
\paragraph{Velocity and score.} FLUX uses $x_t=(1-t)x_0+t\epsilon$ and predicts $v=\mathbb{E}[\epsilon-x_0\,|\,x_t,c]$. Substituting $x_0=(x_t-t\epsilon)/(1-t)$ and Tweedie's formula $\mathbb{E}[\epsilon\,|\,x_t,c]=-t\nabla\log p_t(x_t|c)$,
\begin{equation}
v(x_t;c)=-\frac{x_t+t\,\nabla\log p_t(x_t\,|\,c)}{1-t},
\end{equation}
affine in the score with coefficients that do not depend on the condition $c$. Hence, without the mask, \cref{eq:swap} is the velocity of the score
\begin{equation}
\begin{aligned}
&\nabla\log p_t(x|\Omega)+w\,\nabla\log\frac{p_t(x|\Omega)}{p_t(x|\Omega^{\pi})}\\
&\quad=\nabla\log\Big(p_t(x|\Omega)^{1+w}\,p_t(x|\Omega^{\pi})^{-w}\Big),
\end{aligned}
\end{equation}
the tilt quoted in \cref{sec:swap}. As for classifier-free guidance, this holds per noise level; the guided field is not in general the score of a noised version of a single tilted distribution, so the tilt is an idealization. The mask $M$ restricts the update to region tokens; $M\odot\nabla\ell$ is still an ascent direction of the log ratio $\ell$, since $\langle M\odot\nabla\ell,\nabla\ell\rangle=\|M\odot\nabla\ell\|^2\ge0$.

\paragraph{Which images the tilt favors.} Assume, as an idealization, \emph{binding equivariance}: for congruent regions, the model's density with the subjects bound to the shifted regions equals its density for the image whose region contents are shifted back, $p(x|\Omega^{\pi})=p(\sigma x|\Omega)$, where $\sigma$ moves the content of $\Omega_{\pi(j)}$ to $\Omega_j$. With $f=\log p(\cdot|\Omega)$ the log ratio is $\ell(x)=f(x)-f(\sigma x)$, and
\begin{itemize}
\item if the regions show the same content (two copies, or one hybrid in both), $\sigma x=x$ and $\ell(x)=0$: the tilt gives such images no weight;
\item for a correct composition $x^{\ast}$, $\sigma x^{\ast}$ is the wrongly bound image and $\ell(x^{\ast})>0$ whenever the model prefers the correct binding;
\item over each orbit, $\sum_{q=0}^{m-1}\ell(\sigma^qx)=f(x)-f(\sigma^mx)=0$ since $\sigma^m=\mathrm{id}$: the tilt only moves mass within the orbit $\{x,\sigma x,\dots\}$ towards its correctly bound members, and, after normalization, away from the fixed points of $\sigma$, \ie, from homogenized images.
\end{itemize}
A permutation $\pi$ with a fixed point $j$ would leave region $\Omega_j$ with the correct binding in the negative pass, so $v-v^{\pi}$ would carry little signal there; the cyclic shift has no fixed point for any $m\ge2$, so one negative pass covers every region.

\paragraph{What cancels.} The two passes share $x_i$, the prompt embedding, the gating and the adapter; they differ only in the bias, by $\beta_i-\beta_i^{\pi}$, which is supported on the region--text and region--region entries. To first order, $v-v^{\pi}\approx J_{\beta}\,[\beta_i-\beta_i^{\pi}]$ with $J_{\beta}$ the Jacobian of the network output with respect to the logit bias, so contributions that do not depend on which subject is bound to which region (scene, lighting, $W_0$, the common trigger) cancel. Swapping also the LoRA gating would add a weight-space difference to the negative pass; on the dev seeds the two variants are within noise (\cref{tab:hp_stage2}), and we use the bias-only swap, which keeps the negative pass closest to the positive one.

\section{Hyperparameter sensitivity}
\label{sec:app_hparam}
\begin{table*}[t]
\centering\footnotesize
\setlength{\tabcolsep}{4pt}
\caption{\textbf{Stage-2 sensitivity} (dev seeds 0--5, $n{=}120$ per cell, $K{=}10$ and $K{=}20$). Each block varies one
hyperparameter around the setting in its header; $^\dagger$ = value used by MORE, \textbf{bold} = the final configuration.
``--'' = not run. At $K{=}10$ the paired bootstrap SEs of differences to the final row are .008--.041 for \bothC{} and
.004--.010 for \minC/\meanC.}
\label{tab:hp_stage2}
\begin{tabular}{@{}lcccc|ccc@{}}
\toprule
& \multicolumn{4}{c|}{$K=10$} & \multicolumn{3}{c}{$K=20$} \\
setting & \col$\downarrow$ & \bothC & \minC & \meanC & \bothC & \minC & \meanC \\
\midrule
\textbf{MORE} ($\gamma{=}0.3$, $w{=}2$, ii, bias swap) & \textbf{.008} & \textbf{.942} & \textbf{.218} & \textbf{.349} & \textbf{.742} & \textbf{.248} & \textbf{.364} \\
\midrule
\multicolumn{8}{@{}l}{\textit{gating length $\gamma$} (ii, $w{=}2$)} \\
\quad $\gamma=0$ (ungated) & .025 & .800 & .198 & .350 & .717 & .253 & .378 \\
\quad $\gamma=0.2$ & .008 & .950 & .215 & .348 & .750 & .235 & .358 \\
\multicolumn{8}{@{}l}{\textit{gating length $\gamma$} (ii, $w{=}1$)} \\
\quad all steps & .017 & .917 & .205 & .322 & -- & -- & -- \\
\quad $\gamma=0.3^\dagger$ & .008 & .917 & .204 & .339 & .742 & .228 & .371 \\
\multicolumn{8}{@{}l}{\textit{gating length $\gamma$} (no ii, no guidance)} \\
\quad all steps & .025 & .858 & .185 & .297 & -- & -- & -- \\
\quad $\gamma=0.5$ & .017 & .858 & .188 & .309 & -- & -- & -- \\
\quad $\gamma=0.3^\dagger$ & .017 & .858 & .186 & .317 & -- & -- & -- \\
\midrule
\multicolumn{8}{@{}l}{\textit{swap-guidance weight $w$} ($\gamma{=}0.3$, ii)} \\
\quad $w=0$ & .025 & .850 & .175 & .319 & -- & -- & -- \\
\quad $w=1$ & .008 & .917 & .204 & .339 & .742 & .228 & .371 \\
\quad $w=3$ & .025 & .942 & .211 & .341 & -- & -- & -- \\
\midrule
\multicolumn{8}{@{}l}{\textit{image--image repulsion} ($\gamma{=}0.3$, $w{=}1$)} \\
\quad none & .008 & .892 & .190 & .320 & -- & -- & -- \\
\quad $-p_i/2^\dagger$ & .008 & .917 & .204 & .339 & .742 & .228 & .371 \\
\multicolumn{8}{@{}l}{\textit{image--image repulsion} (ungated, no guidance)} \\
\quad none & .025 & .675 & .157 & .323 & -- & -- & -- \\
\quad $-p_i/2^\dagger$ & .008 & .750 & .170 & .334 & -- & -- & -- \\
\quad $-p_i$ & .067 & .742 & .202 & .346 & -- & -- & -- \\
\quad $-1.5\,p_i$ & .333 & .500 & .160 & .257 & -- & -- & -- \\
\midrule
\multicolumn{8}{@{}l}{\textit{what the guidance negative swaps} (gating all steps, ii, $w{=}1$)} \\
\quad bias and LoRA gating & .008 & .933 & .204 & .323 & -- & -- & -- \\
\quad bias only$^\dagger$ & .017 & .917 & .205 & .322 & -- & -- & -- \\
\bottomrule
\end{tabular}
\end{table*}

\begin{table*}[t]
\centering\footnotesize
\begin{minipage}[t]{0.49\textwidth}
\centering
\setlength{\tabcolsep}{3.2pt}
\caption{\textbf{Attention-bias design} (dev seeds 0--5, $n{=}120$ per row, $K{=}10$ unless marked). The rows use the
ungated merge without repulsion or guidance (the first design round), so the absolute numbers are below
\cref{tab:hp_stage2}. $^\dagger$ = value used by MORE ($p_0{=}2$, cosine decay, no blocking, bias over all steps).
$^\ast$ bias only over the first 30\% of the steps.}
\label{tab:hp_bias}
\begin{tabular}{@{}lcccc@{}}
\toprule
setting & \col$\downarrow$ & \bothC & \minC & \meanC \\
\midrule
static merge (no attention edit) & .008 & .275 & .096 & .271 \\
\midrule
\multicolumn{5}{@{}l}{\textit{blocking / own-text bias} (constant schedule)} \\
\quad block $-\infty$, no bias$^\ast$ & .033 & .133 & .069 & .232 \\
\quad block $-4$, no bias & .025 & .075 & .065 & .225 \\
\quad block $-4$, $p_0{=}2$ & .017 & .625 & .155 & .324 \\
\quad no blocking, $p_0{=}2^\dagger$ & .017 & .675 & .156 & .321 \\
\quad no layout, $+2$ to all subject text & .042 & .492 & .137 & .294 \\
\midrule
\multicolumn{5}{@{}l}{\textit{bias strength $p_0$}, constant schedule} \\
\quad $p_0=1.5$ & .017 & .592 & .153 & .319 \\
\quad $p_0=2^\dagger$ & .017 & .675 & .156 & .321 \\
\quad $p_0=2.5$ & .008 & .650 & .150 & .322 \\
\quad $p_0=3$ & .167 & .375 & .087 & .193 \\
\multicolumn{5}{@{}l}{\textit{bias strength $p_0$}, cosine decay} \\
\quad $p_0=2^\dagger$ & .025 & .675 & .157 & .323 \\
\quad $p_0=2.5$ & .025 & .683 & .148 & .322 \\
\quad $p_0=3$ & .033 & .733 & .161 & .309 \\
\multicolumn{5}{@{}l}{\textit{bias strength $p_0$}, cosine decay, layout gating at all steps} \\
\quad $p_0=2^\dagger$ & .025 & .858 & .185 & .297 \\
\quad $p_0=3$ & .025 & .758 & .156 & .250 \\
\midrule
\multicolumn{5}{@{}l}{\textit{decay}, layout gating at all steps, $p_0{=}2$} \\
\quad constant, $K{=}10$ & .017 & .867 & .187 & .298 \\
\quad cosine$^\dagger$, $K{=}10$ & .025 & .858 & .185 & .297 \\
\quad constant, $K{=}20$ & .042 & .567 & .182 & .321 \\
\quad cosine$^\dagger$, $K{=}20$ & .008 & .625 & .190 & .337 \\
\midrule
\multicolumn{5}{@{}l}{\textit{bias window} (fraction of steps, no blocking, $p_0{=}2$, constant)} \\
\quad 0.3 & .017 & .600 & .150 & .322 \\
\quad 0.5 & .017 & .667 & .153 & .324 \\
\quad 1$^\dagger$ & .017 & .675 & .156 & .321 \\
\bottomrule
\end{tabular}
\end{minipage}\hfill
\begin{minipage}[t]{0.47\textwidth}
\centering
\setlength{\tabcolsep}{4pt}
\caption{\textbf{Stage-1 sensitivity} on the single-subject benchmark: mean DINOv2 over three dev seed pairs
$\{0,1\},\{2,3\},\{4,5\}$ ($n{=}80$ images per pair at $K{=}10$, 160 at $K{=}20$); blocks or columns marked $^\ast$ are
screens on seeds $\{0,1\}$ only and are not directly comparable with the three-pair means (at $K{=}10$ the $\{0,1\}$
values are about .01 higher).
Each block varies one factor at the base in its header. \textbf{Bold} = value used by MORE.
Paired differences over (seed pair, subject) cells, mean $\pm$ SE, at $K{=}10$ / $K{=}20$: mixing weight 3 vs.\ 0
$+.029{\pm}.009$ / $+.027{\pm}.007$; de-crowding on vs.\ off $+.025{\pm}.006$ / $+.026{\pm}.005$;
$g{=}1$ vs.\ $0.8$ ($K{=}10$) $+.025{\pm}.007$; $\rho{=}0.05$ vs.\ $0.1$ ($K{=}10$) $+.011{\pm}.007$; $r{=}16$ vs.\ $8$ ($K{=}10$)
$-.030{\pm}.010$.}
\label{tab:hp_stage1}
\begin{tabular}{@{}lcc@{}}
\toprule
setting & $K=10$ & $K=20$ \\
\midrule
\multicolumn{3}{@{}l}{\textit{rank $r$}$^\ast$ ($K{=}10$: $\rho{=}0.2$, $g{=}0.8$; $K{=}20$: $\rho{=}0.1$, $g{=}1$)} \\
\quad 4 & .429 & -- \\
\quad 6 & .468 & -- \\
\quad 8 & \textbf{.482} & .358 \\
\quad 12 & .435 & .408 \\
\quad 16 & .436 & \textbf{.441} \\
\quad 20 & -- & .373 \\
\quad 32 & .431 & -- \\
\midrule
\multicolumn{3}{@{}l}{\textit{relative ridge $\rho$} ($r{=}8$ / $16$, $g{=}1$; $K{=}20$ column$^\ast$)} \\
\quad 0.2 & .495 & .421 \\
\quad 0.1 & .496 & \textbf{.441} \\
\quad 0.05 & \textbf{.507} & .443 \\
\midrule
\multicolumn{3}{@{}l}{\textit{energy gain $g$}$^\ast$ ($K{=}10$: $r{=}8$, $\rho{=}0.2$; $K{=}20$: $r{=}16$, $\rho{=}0.1$)} \\
\quad 0.6 & .443 & -- \\
\quad 0.8 & .482 & -- \\
\quad 1 & \textbf{.510} & \textbf{.441} \\
\quad 1.2 & .501 & .380 \\
\quad 1.5 & .188 & -- \\
\midrule
\multicolumn{3}{@{}l}{\textit{calibration mixing weight} ($r{=}8$, $\rho{=}0.2$, $g{=}0.8$)} \\
\quad 0 (noise latents only) & .441 & .348 \\
\quad 1 & .462 & .367 \\
\quad 3 & \textbf{.469} & \textbf{.375} \\
\midrule
\multicolumn{3}{@{}l}{\textit{de-crowding} (final merge: $r{=}8$/$16$, $\rho{=}0.05$/$0.1$, $g{=}1$)} \\
\quad off & .507 & .445 \\
\quad on & \textbf{.532} & \textbf{.471} \\
\quad control: random directions & .509 & .446 \\
\bottomrule
\end{tabular}
\end{minipage}
\end{table*}
All hyperparameters were set on development seeds 0--5 and never on the held-out seeds 6--11 that the main tables report.
\Cref{tab:hp_stage2,tab:hp_bias} vary one Stage-2 hyperparameter at a time on the pair benchmark (20 prompts $\times$
seeds 0--5, $n{=}120$ per row). The Stage-1 hyperparameters (rank, ridge, energy gain, calibration mixing weight and
de-crowding) were chosen on a single-subject benchmark on the same seeds 0--5: one subject per prompt, all $K$ subjects,
4 prompts each, scored by DINOv2 similarity to the references (\cref{tab:hp_stage1}). This benchmark has no collage-aware
variant. The guided-step set $\{0,2,4,6,8\}$ was not tuned.

\paragraph{Findings.} (i) \emph{Gating length.} At $K{=}10$, gating the first $20\%$ or $30\%$ of the steps is equivalent and both clearly beat no gating on \bothC{} ($.95$/$.94$ vs.\ $.80$); gating all steps lowers \meanC{}, which is why gating stops early. At $K{=}20$ the ungated variant is tied with the final configuration on \bothC{} and nominally better on \minC/\meanC{}, and the held-out $K{=}10$ ablation shows the same trade-off (gating: $+.12{\pm}.04$ \bothC{}, $-.02{\pm}.01$ \meanC{}); gating buys coexistence, not identity. (ii) \emph{Guidance weight.} $w{=}1,2,3$ all improve over $w{=}0$; $w{=}2$ and $w{=}3$ give the same \bothC{} at $K{=}10$, so the optimum is shallow, and at $K{=}20$ $w{=}2$ raises the collage rate from $.008$ to $.058$. (iii) \emph{Repulsion and swap target.} $-p_i/2$ helps slightly; $-1.5p_i$ causes collages. Swapping the bias and the LoRA gating is within noise of the bias-only swap; we keep the latter (\cref{sec:app_swap}). (iv) \emph{Bias.} Blocking without a positive bias is harmful and the $+p$ own-text bias is the active ingredient; $p_0\in[2,2.5]$ is flat and $p_0{=}3$ fails with a constant schedule, which the cosine decay prevents; at $K{=}20$ the cosine decay is better ($.625$ vs.\ $.567$). (v) \emph{Stage 1.} The single-subject metric peaks at $r{=}8$ ($K{=}10$) and $r{=}16$ ($K{=}20$) and at $g{=}1$, \ie, matching the own-LoRA energy; at $K{=}20$, $\rho{=}0.05$ and $0.1$ are tied ($.443$ vs.\ $.441$, screened on seeds $\{0,1\}$) and we keep $0.1$; $\rho{=}0.05$ at $K{=}10$ and the mixing weight $3$ lie at the edge of the tested grid, so we claim only that they are best \emph{within} it. De-crowding helps on the single-subject benchmark ($+.025{\pm}.006$) and its random-direction control does not, but on the held-out pair benchmark removing it from MORE is tied or nominally better (\cref{tab:ablation}), so its single-subject gain does not carry over to composition.

\section{Unselected qualitative results}
\label{sec:app_unsel}
\begin{figure*}[p]
\centering
\includegraphics[width=\textwidth,height=0.84\textheight,keepaspectratio]{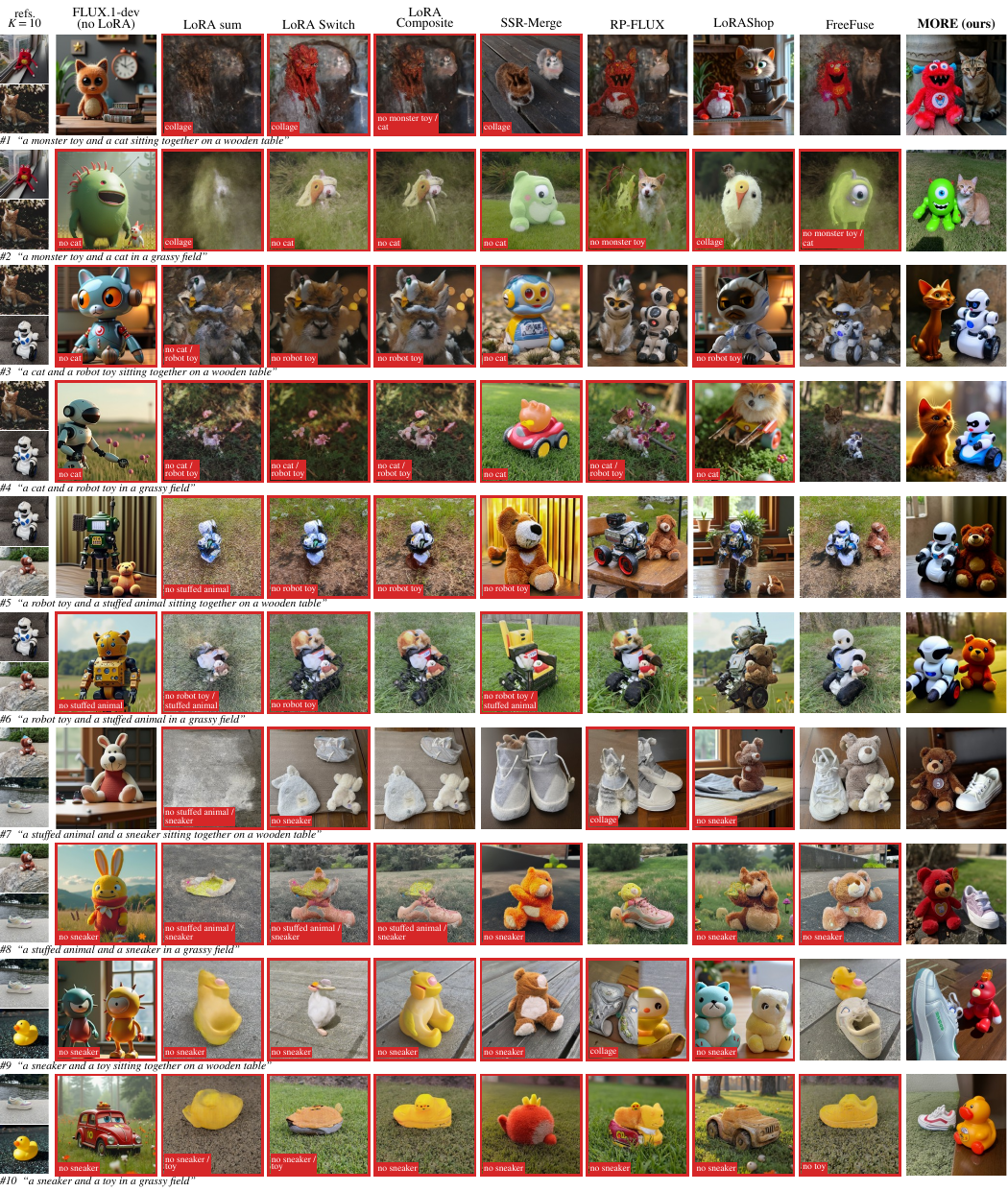}
\caption{\textbf{Unselected comparison, $K{=}10$ pool} (prompts 1--10; 11--20 on the next page). All 20 prompts of the pair benchmark at one fixed held-out seed (seed 6, the first held-out seed), ordered by prompt index; no image, prompt or seed was selected. Every method uses the same prompt and the same seed (seed index 6, \cref{sec:setup}). Columns as in \cref{tab:main}: FLUX.1-dev without LoRA, LoRA sum, LoRA Switch, LoRA Composite, SSR-Merge (published calibration, whole pool), RP-FLUX (LoRA weight 0.5, oracle layout), LoRAShop, FreeFuse (both automatic masks), and MORE (whole pool, oracle left/right layout). A \textcolor{red}{red frame} marks an image that fails the automatic coexistence test of \bothC{}: a subject not detected by OWLv2 (score ${<}0.2$; tag ``no $\langle$class$\rangle$'') or a CLIP-flagged collage (tag ``collage''). The frames are produced by the metric, not by us; they are not a judgement of identity, which \bothC{} does not measure (an unframed image may still show a wrong identity or a hybrid). Flagged images at this seed (of 20): FLUX.1-dev (no LoRA) 16, LoRA sum 20, LoRA Switch 18, LoRA Composite 17, SSR-Merge 19, RP-FLUX 12, LoRAShop 15, FreeFuse 8, MORE (ours) 0. For reference, over the six held-out seeds 6--11 MORE passes 17--20 of 20 per seed (seed 6: 20; \bothC{} 0.942 in \cref{tab:main}) and FreeFuse 8--13 (seed 6: 12; 0.508).}
\label{fig:supp_pairs_K10}
\end{figure*}

\begin{figure*}[p]
\centering
\includegraphics[width=\textwidth,height=0.84\textheight,keepaspectratio]{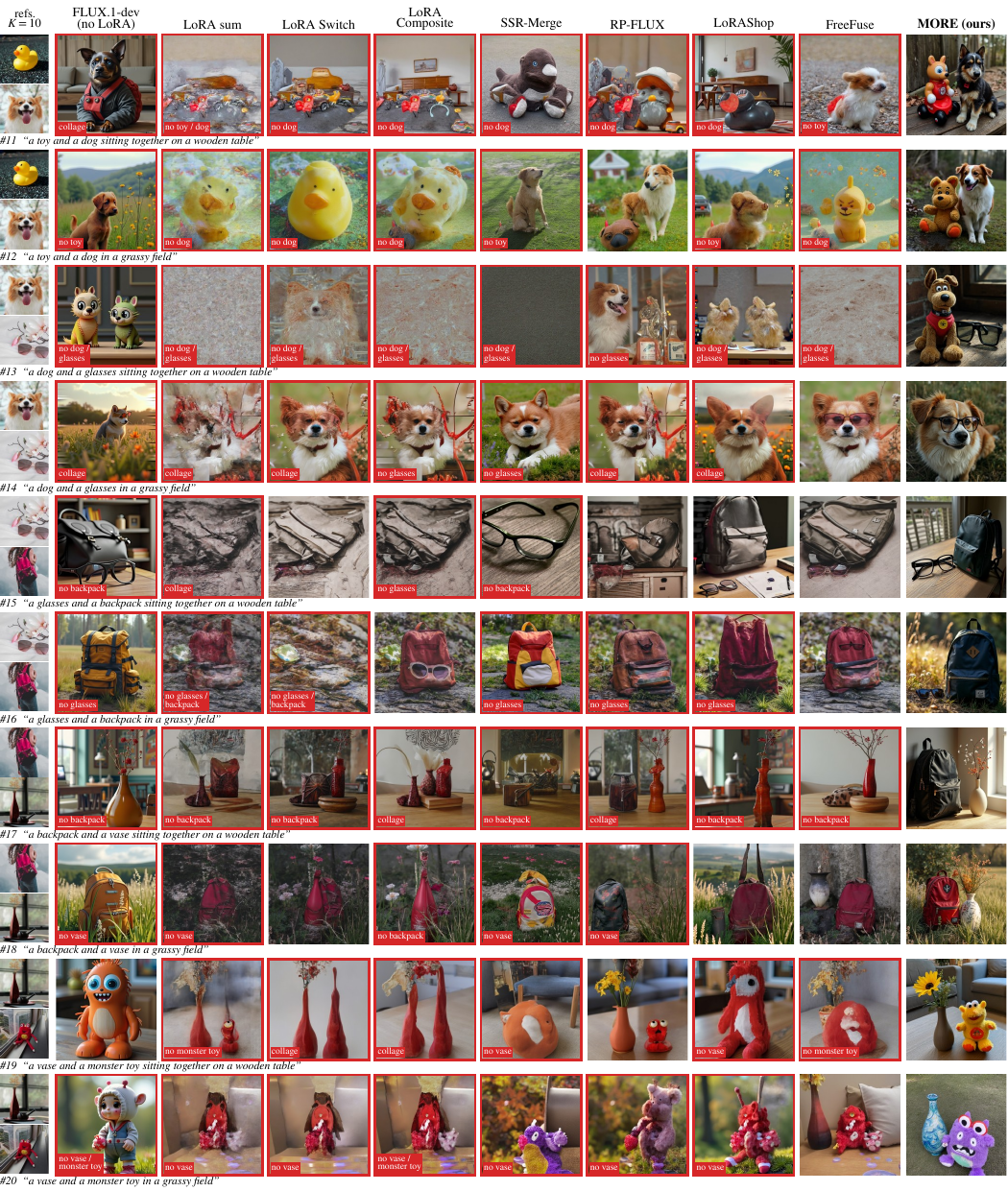}
\caption{\textbf{Unselected comparison, $K{=}10$ pool, continued} (prompts 11--20); see \cref{fig:supp_pairs_K10}.}
\label{fig:supp_pairs_K10b}
\end{figure*}

\begin{figure*}[p]
\centering
\includegraphics[width=\textwidth,height=0.84\textheight,keepaspectratio]{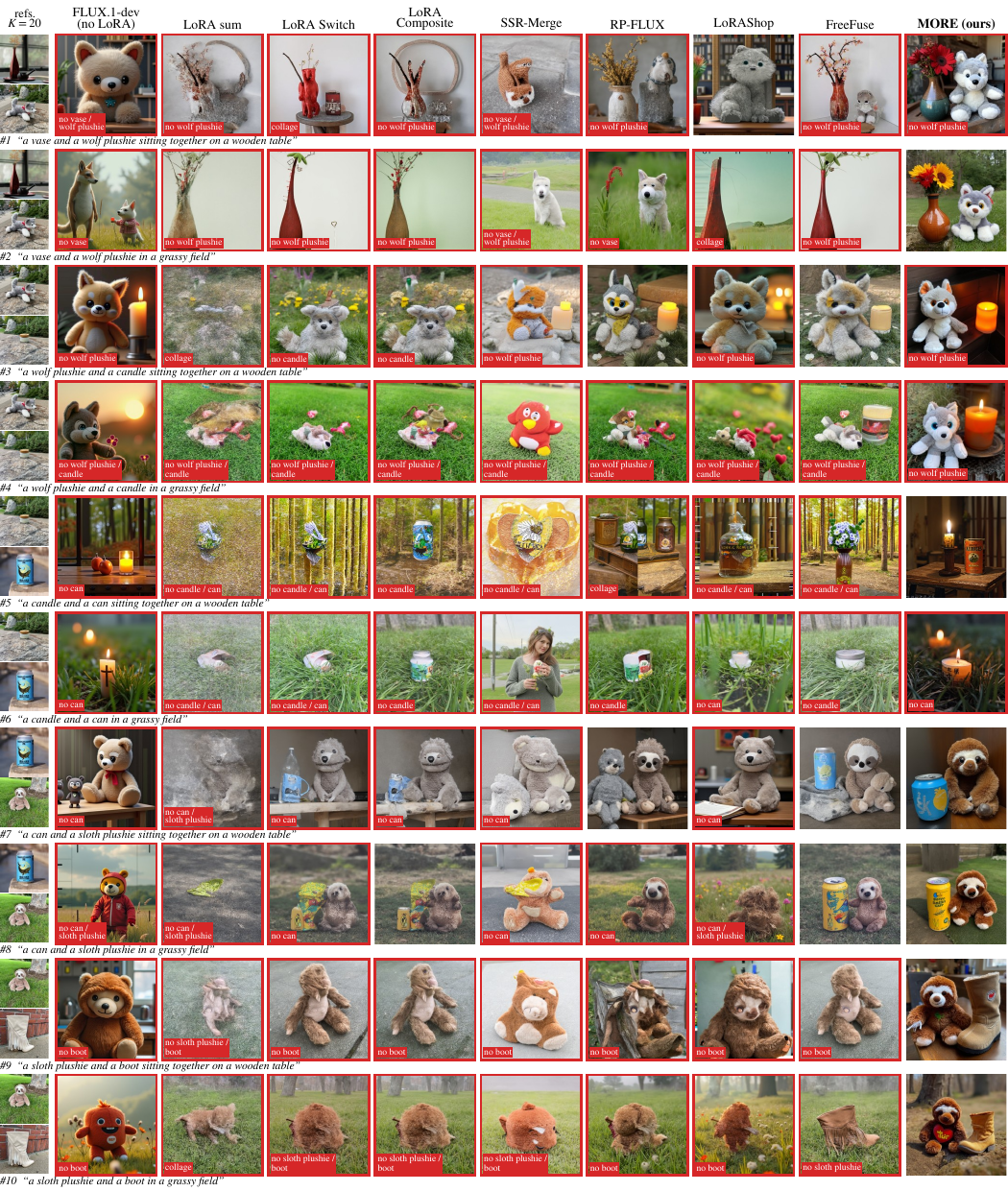}
\caption{\textbf{Unselected comparison, $K{=}20$ pool} (prompts 1--10; 11--20 on the next page). All 20 prompts of the pair benchmark at one fixed held-out seed (seed 6, the first held-out seed; prompts and subject pairs of the $K{=}20$ benchmark), ordered by prompt index; no image, prompt or seed was selected. Every method uses the same prompt and the same seed (seed index 6, \cref{sec:setup}). Columns as in \cref{tab:main}: FLUX.1-dev without LoRA, LoRA sum, LoRA Switch, LoRA Composite, SSR-Merge (published calibration, whole pool), RP-FLUX (LoRA weight 0.5, oracle layout), LoRAShop, FreeFuse (both automatic masks), and MORE (whole pool, oracle left/right layout). A \textcolor{red}{red frame} marks an image that fails the automatic coexistence test of \bothC{}: a subject not detected by OWLv2 (score ${<}0.2$; tag ``no $\langle$class$\rangle$'') or a CLIP-flagged collage (tag ``collage''). The frames are produced by the metric, not by us; they are not a judgement of identity, which \bothC{} does not measure (an unframed image may still show a wrong identity or a hybrid). Flagged images at this seed (of 20): FLUX.1-dev (no LoRA) 16, LoRA sum 19, LoRA Switch 18, LoRA Composite 17, SSR-Merge 19, RP-FLUX 13, LoRAShop 13, FreeFuse 11, MORE (ours) 4. For reference, over the six held-out seeds 6--11 MORE passes 13--17 of 20 per seed (seed 6: 16; \bothC{} 0.783 in \cref{tab:main}) and FreeFuse 4--9 (seed 6: 9; 0.358).}
\label{fig:supp_pairs_K20}
\end{figure*}

\begin{figure*}[p]
\centering
\includegraphics[width=\textwidth,height=0.84\textheight,keepaspectratio]{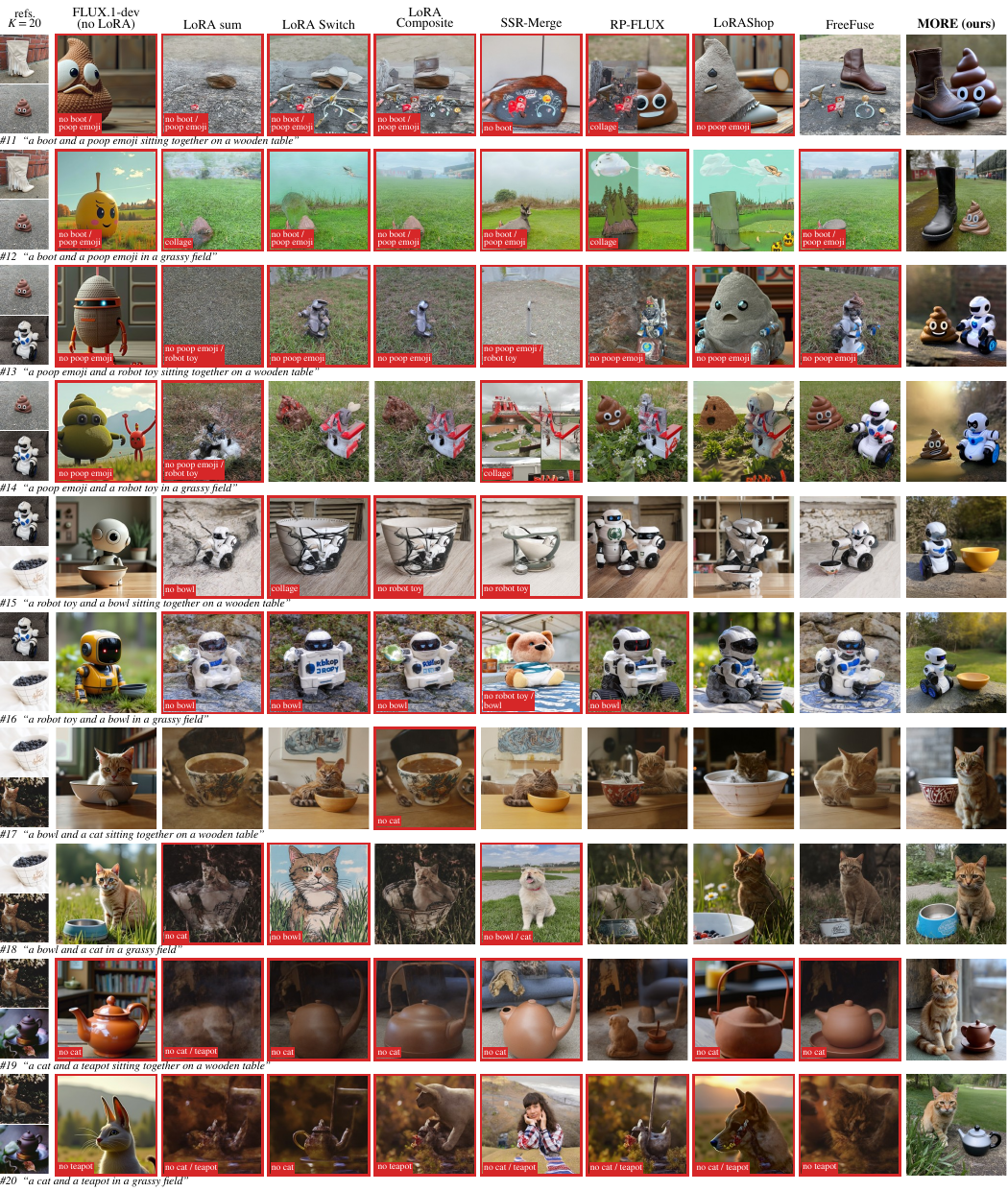}
\caption{\textbf{Unselected comparison, $K{=}20$ pool, continued} (prompts 11--20); see \cref{fig:supp_pairs_K20}.}
\label{fig:supp_pairs_K20b}
\end{figure*}

\begin{figure*}[p]
\centering
\includegraphics[width=\textwidth,height=0.8\textheight,keepaspectratio]{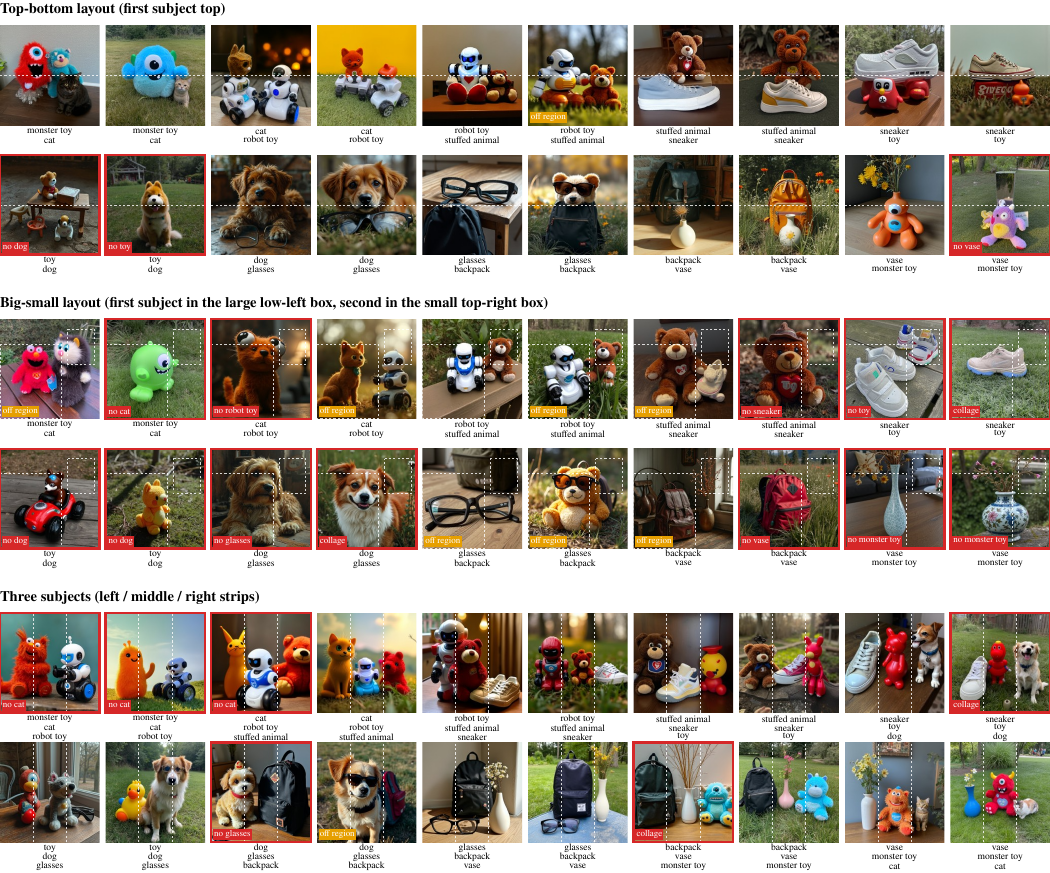}
\caption{\textbf{Unselected layout and three-subject results of MORE} ($K{=}10$ pool, held-out seed 6, all 20 prompts per setting in prompt order, no selection). Top: top-bottom layout; middle: big-small layout (large low-left box, small top-right box); bottom: three subjects in left/middle/right strips. Dashed lines: the layout regions; the subject names under each image are listed in region order. A \textcolor{red}{red frame} marks a failure of the criterion of \cref{tab:general} (\bothC{} / \allC{}): a subject not detected by OWLv2 (score ${<}0.2$) or a CLIP-flagged collage. An \textcolor{orange}{amber tag} marks an image in which all subjects are detected but at least one box centre lies outside its region (counted by \bothC{} but not by ``hit''). Failures: 3/20 top-bottom, 12/20 big-small, 6/20 three subjects; off-region: 1, 7, 1.}
\label{fig:supp_layouts}
\end{figure*}

The qualitative figures of the main text show held-out images of our method ranked automatically by detection score and DINOv2 identity among those that pass the coexistence test, \ie, our best cases. To show typical behavior instead, \cref{fig:supp_pairs_K10,fig:supp_pairs_K10b,fig:supp_pairs_K20,fig:supp_pairs_K20b} show \emph{all} 20 prompts of the pair benchmark at the first held-out seed (seed 6) for every method of \cref{tab:main}, and \cref{fig:supp_layouts} shows all 20 prompts of each generalization setting of \cref{tab:general} for our method. Nothing was selected: prompts appear in benchmark order and the seed is fixed in advance. Red frames are computed by the same detector and collage test as the tables, not by us; seed 6 happens to be a favorable seed for our method at $K{=}10$ (20/20 pass, against 17--20 over seeds 6--11), which the captions state together with the per-seed ranges.

\clearpage

\clearpage
\section{Identity confusion}
\label{sec:app_conf}
\Cref{tab:conf} reports a slot-confusion analysis in the spirit of MuDI~\cite{mudi}: for each detected crop $k$ we compute DINOv2 similarity to the references of every prompt subject $j$; ``own'' is the mean diagonal, ``cross'' the mean off-diagonal, ``conf'' the fraction of crops closer to another prompt subject than to their own, and ``crop--crop'' the cosine between the two crops. Our crops have the highest own \emph{and} cross similarity and the highest confusion. This is partly a confound: when a baseline drops a subject, the second crop is a wrong object or background, which lowers cross similarity and confusion; and our two crops are often similar plush or toy objects under identical lighting. It nevertheless shows that our pipeline does not remove identity leakage between the two regions, and that swap guidance, which lowers feature similarity (\cref{fig:mechanism}), does not lower crop similarity.

\begin{table*}[h]
\centering
\footnotesize
\setlength{\tabcolsep}{3.5pt}
\caption{Identity confusion (held-out, $n{=}120$; DINOv2 on OWLv2 crops). Higher ``own'' is better; the other columns are confounded by missing subjects (see text).}
\label{tab:conf}
\begin{tabular}{@{}lcccc|cccc@{}}
\toprule
& \multicolumn{4}{c|}{$K=10$} & \multicolumn{4}{c}{$K=20$} \\
& own & cross & conf & crop--crop & own & cross & conf & crop--crop \\
\midrule
LoRA sum & .100 & .080 & .122 & .675 & .098 & .071 & .163 & .583 \\
SSR-Merge & .192 & .120 & .093 & .447 & .148 & .099 & .146 & .427 \\
RP-FLUX & .280 & .197 & .174 & .545 & .279 & .167 & .173 & .450 \\
LoRAShop & .240 & .155 & .150 & .469 & .228 & .131 & .170 & .427 \\
FreeFuse & .319 & .204 & .192 & .545 & .332 & .207 & .291 & .486 \\
our merge, static & .277 & .181 & .110 & .500 & .321 & .175 & .137 & .420 \\
gating + bias & .296 & .264 & .408 & .692 & .351 & .253 & .348 & .622 \\
MORE & .343 & .297 & .421 & .719 & .386 & .255 & .371 & .567 \\
\midrule
\multicolumn{9}{@{}l}{\textit{dev seeds 0--5, $K{=}10$}} \\
MORE w/o swap guidance & .324 & .263 & .378 & .657 & & & & \\
MORE & .350 & .275 & .379 & .680 & & & & \\
\bottomrule
\end{tabular}
\end{table*}

\section{User study: all 50 trials}
\label{sec:app_us}
\begin{figure*}[p]
\centering
\includegraphics[width=\textwidth,height=0.9\textheight,keepaspectratio]{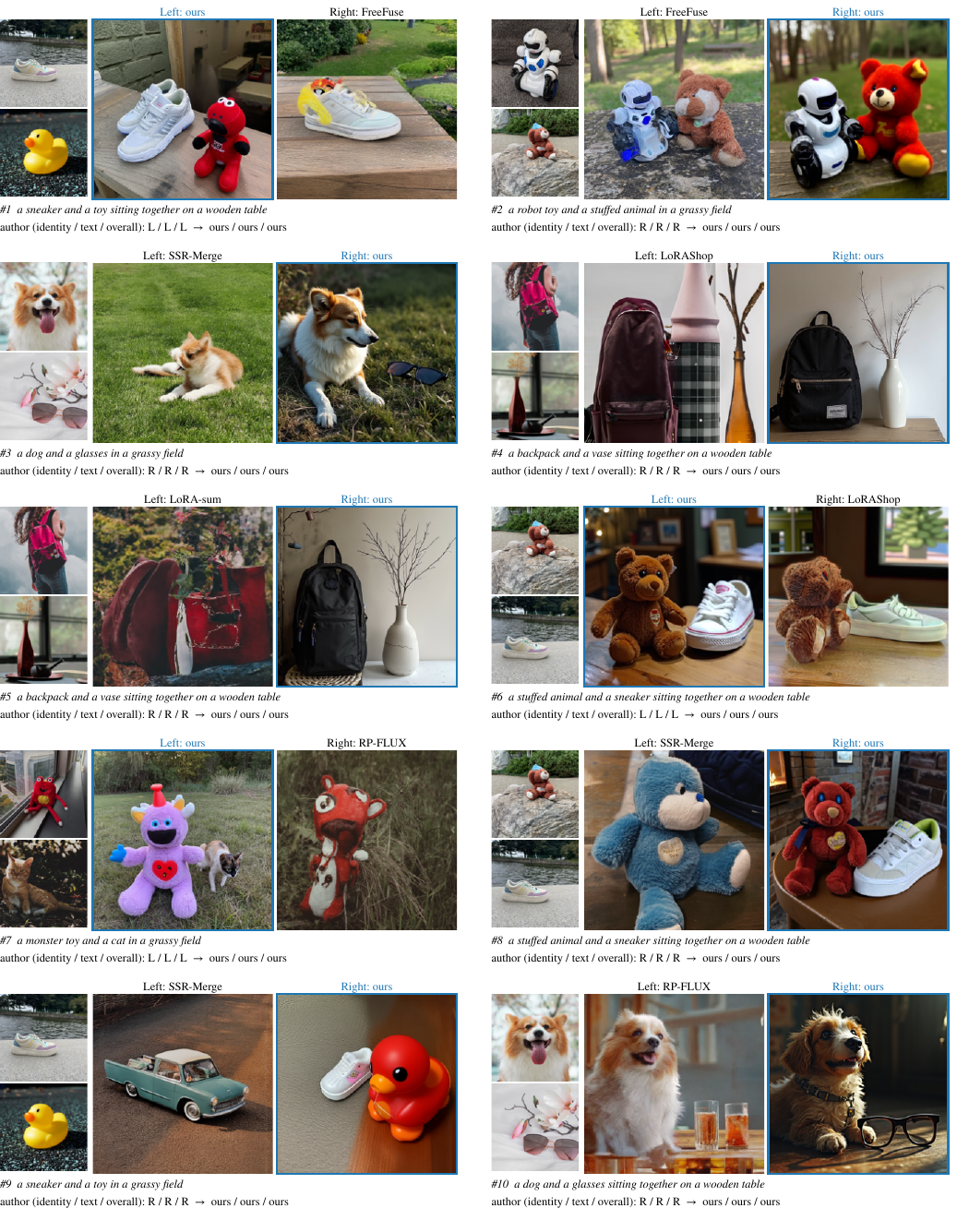}
\caption{User study, trials 1--10 (page 1 of 5). Blue: our image (not shown to the rater).}
\label{fig:us1}
\end{figure*}

\begin{figure*}[p]
\centering
\includegraphics[width=\textwidth,height=0.9\textheight,keepaspectratio]{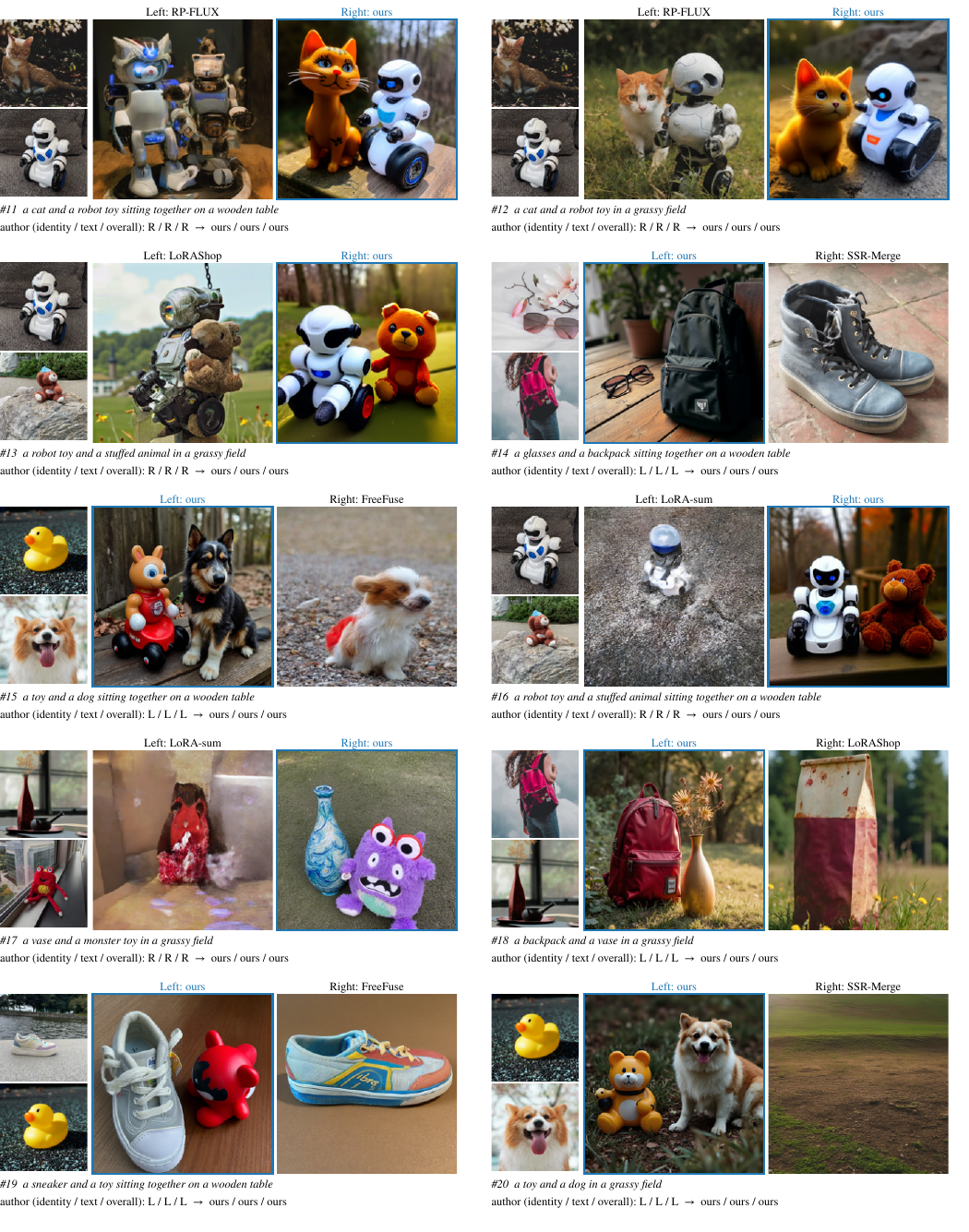}
\caption{User study, trials 11--20 (page 2 of 5). Blue: our image (not shown to the rater).}
\label{fig:us2}
\end{figure*}

\begin{figure*}[p]
\centering
\includegraphics[width=\textwidth,height=0.9\textheight,keepaspectratio]{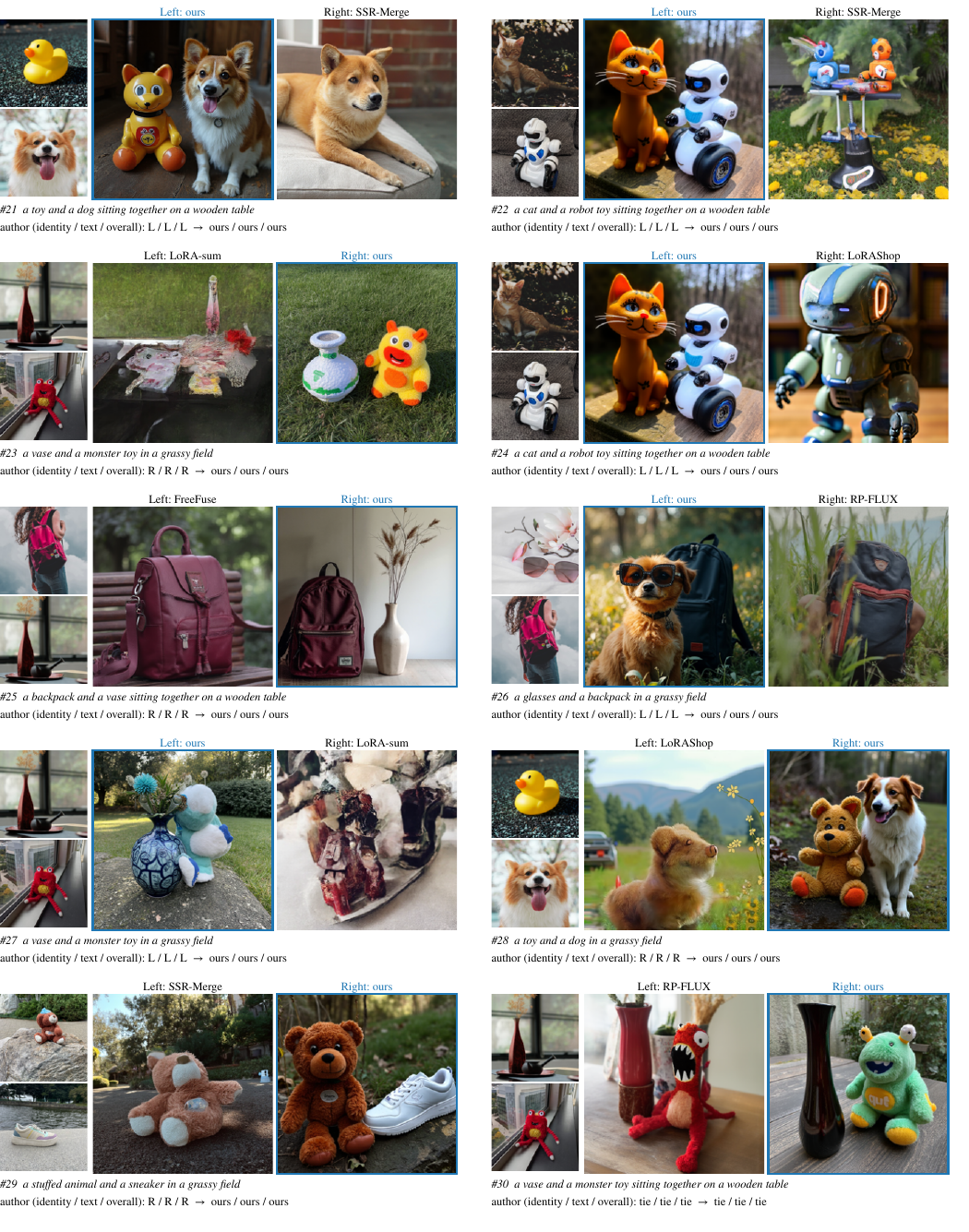}
\caption{User study, trials 21--30 (page 3 of 5). Blue: our image (not shown to the rater).}
\label{fig:us3}
\end{figure*}

\begin{figure*}[p]
\centering
\includegraphics[width=\textwidth,height=0.9\textheight,keepaspectratio]{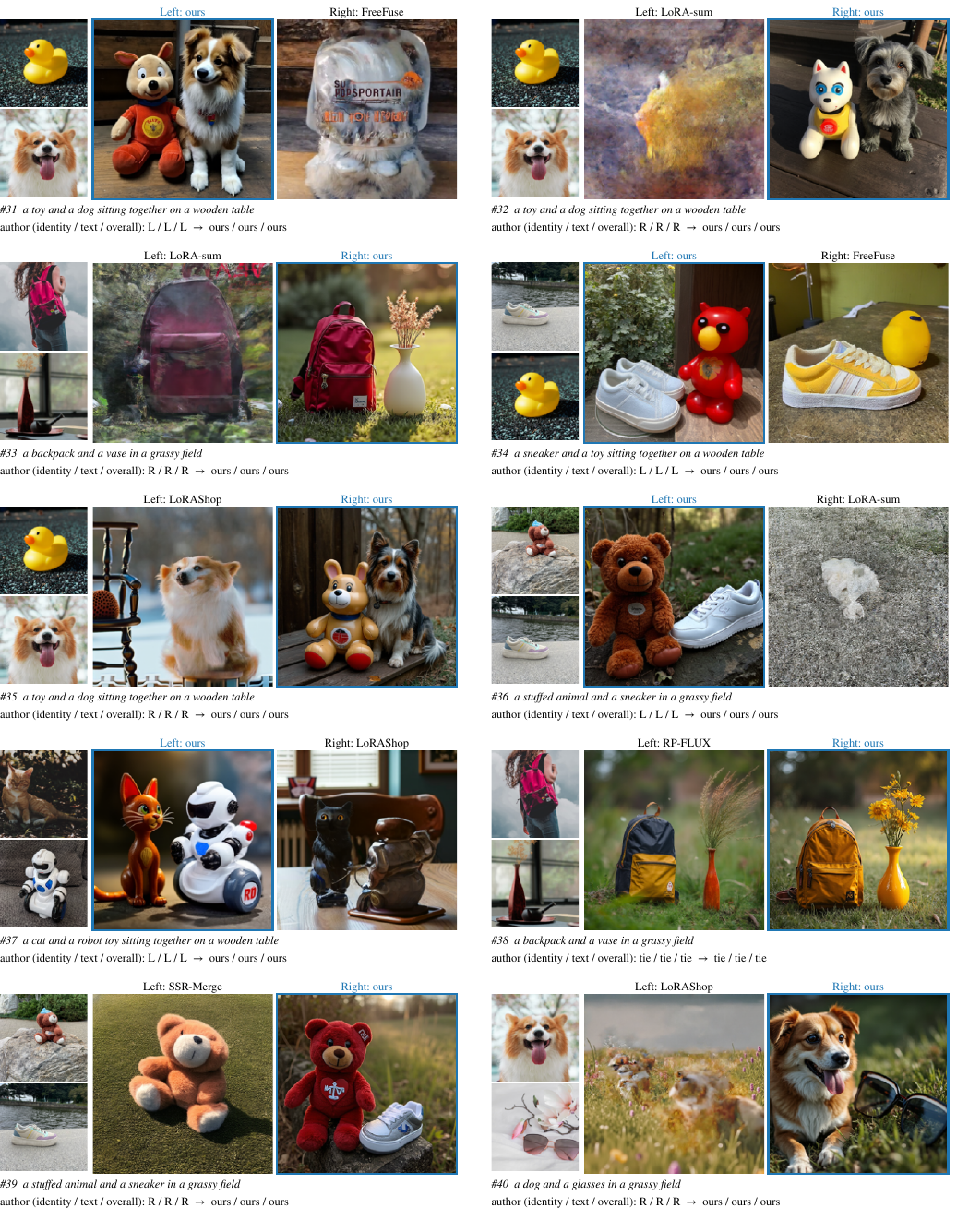}
\caption{User study, trials 31--40 (page 4 of 5). Blue: our image (not shown to the rater).}
\label{fig:us4}
\end{figure*}

\begin{figure*}[p]
\centering
\includegraphics[width=\textwidth,height=0.9\textheight,keepaspectratio]{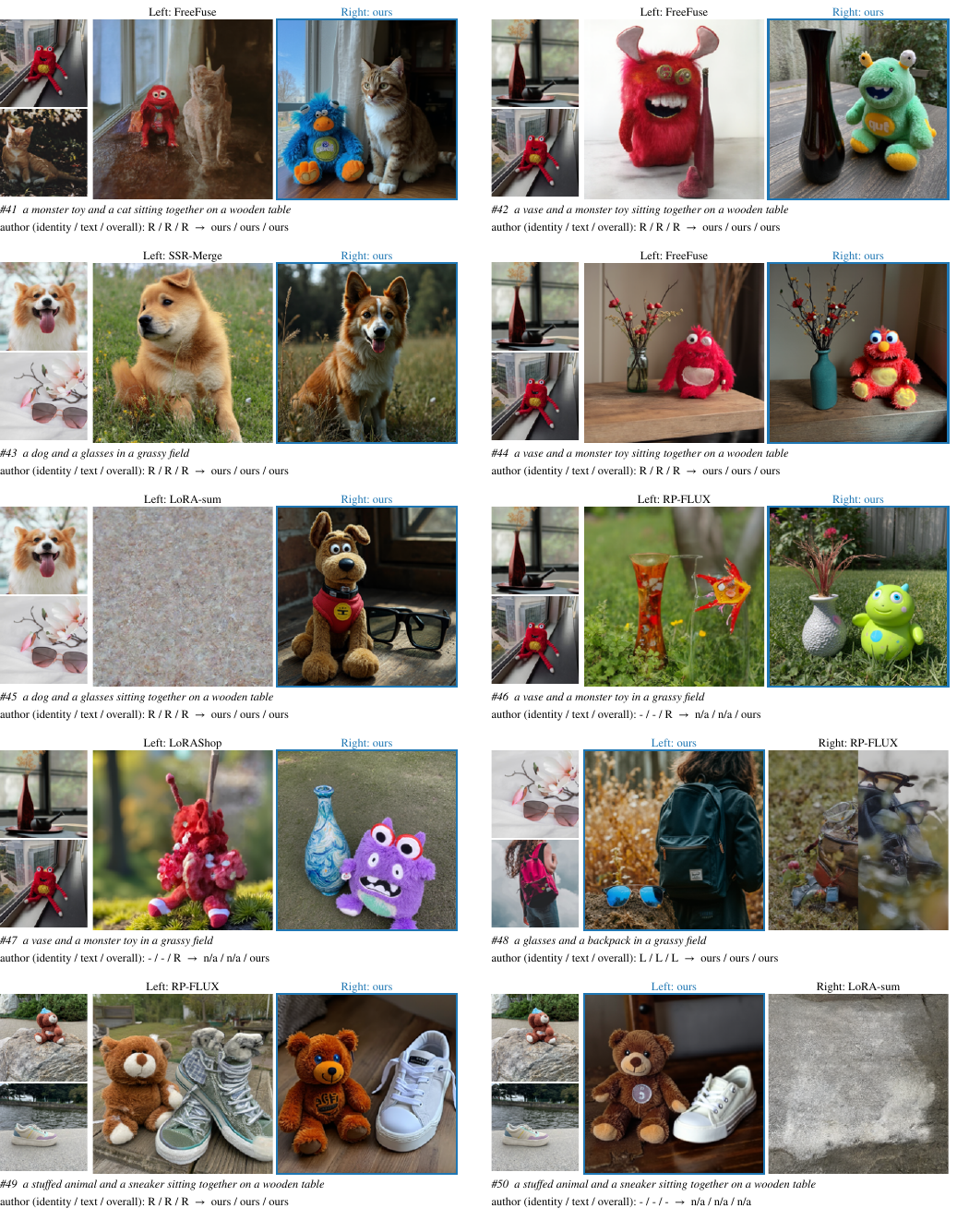}
\caption{User study, trials 41--50 (page 5 of 5). Blue: our image (not shown to the rater).}
\label{fig:us5}
\end{figure*}

The blind study showed, per trial, the two references, the prompt without \sks{}, and two images in random left/right order, with three questions: which image keeps \emph{both} identities, which matches the text better, and which is better overall (left / right / tie). Images were drawn at random from the held-out $K{=}10$ outputs, ten trials per baseline. \Cref{fig:us1,fig:us2,fig:us3,fig:us4,fig:us5} show all 50 trials exactly as rated; the method names and the blue frame around our image were \emph{not} shown to the rater and are added here. Below each trial we give the author's three answers and their meaning (``ours'', ``baseline'', ``tie'', ``n/a'' for a skipped question). Per baseline (identity / text / overall, wins:ties:losses), FreeFuse 10:0:0 / 10:0:0 / 10:0:0; SSR-Merge 10:0:0 / 10:0:0 / 10:0:0; LoRAShop 9:0:0 / 9:0:0 / 10:0:0; LoRA sum 9:0:0 / 9:0:0 / 9:0:0; RP-FLUX 7:2:0 / 7:2:0 / 8:2:0. Seven answers were left blank and are excluded. The rater is the author and the study is no substitute for an independent crowd study; we include it so that readers can check the judgments against the images.


\end{document}